\pdfoutput=1
\documentclass[11pt]{article}

\usepackage[final]{acl}

\usepackage{times}
\usepackage{latexsym}
\usepackage{algorithm}
\usepackage{url}
\usepackage{soul}
\usepackage{booktabs}
\usepackage{array,multirow,graphicx}
\usepackage[normalem]{ulem}
\useunder{\uline}{\ul}{}
\usepackage{todonotes}
\usepackage{amssymb}
\usepackage{stfloats}

\newcommand{\WRP}{\par\qquad\(\hookrightarrow\)\enspace}
\usepackage{verbatim}
\usepackage[bb=boondox]{mathalfa}
\usepackage{subfigure}

\usepackage[noend]{algpseudocode}

\usepackage[T1]{fontenc}
\usepackage[utf8]{inputenc}

\usepackage{microtype}

\usepackage{enumitem}
\usepackage{tikz}
\usepackage{xcolor}
\usepackage{color,collcell}
\usepackage{xstring}

\newcommand*{\MinNumber}{0.0}%
\newcommand*{\MidNumber}{50.0} %
\newcommand*{\MaxNumber}{100.0}%

\newcommand{\ApplyGradient}[1]{%
    \IfDecimal{#1}{
        \ifdim #1 pt > \MidNumber pt
            \pgfmathsetmacro{\PercentColor}{max(min(100.0*(#1 - \MidNumber)/(\MaxNumber-\MidNumber),100.0),0.00)}
            \hspace{-0.33em}\colorbox{green!\PercentColor!yellow}{#1}
        \else
            \pgfmathsetmacro{\PercentColor}{max(min(100.0*(\MidNumber - #1)/(\MidNumber-\MinNumber),100.0),0.00)}
            \hspace{-0.33em}\colorbox{red!\PercentColor!yellow}{#1}
        \fi
    }{#1}
}

\newcolumntype{R}{>{\collectcell\ApplyGradient}c<{\endcollectcell}}

\title{Visual Semantic Parsing: From Images to Abstract Meaning Representation}

\author{Mohamed A. Abdelsalam\textsuperscript{1}, Zhan Shi\textsuperscript{1,2*}, Federico Fancellu\textsuperscript{3\dag}, Kalliopi Basioti\textsuperscript{1,4*}, \\ {\bf Dhaivat J. Bhatt\textsuperscript{1}, Vladimir Pavlovic\textsuperscript{1,4}, Afsaneh Fazly\textsuperscript{1}} \\
\textsuperscript{1}Samsung AI Centre - Toronto\textbf{\ ,\ }\textsuperscript{2}Queen's University\textbf{\ ,\ }\textsuperscript{3}3M\textbf{\ ,\ }\textsuperscript{4}Rutgers University \\
\texttt{\{m.abdelsalam, d.bhatt, a.fazly\}@samsung.com}\textbf{\ ,\ } \texttt{z.shi@queensu.ca} \\ \texttt{f.fancellu0@gmail.com}\textbf{\ ,\ } \texttt{\{kalliopi.basioti, vladimir\}@rutgers.edu} \\\
}

\begin{document}
\maketitle

\begin{comment} % OLD ABSTRACT
\begin{abstract}
Understanding a visual scene is key to the development of a multi-modal interactive system, e.g., language-based visual search or visual question answering. 
Current approaches to scene understanding build visual scene graphs or semantic frames from entities (people/objects), activities, and their relations.
These approaches not only require expensive manual annotation in the form of images paired with their scene graphs or frames, but are also limited in the nature of entities and relations they capture. 
We instead propose to draw on a well-known linguistically-motivated formalism, the Abstract Meaning Representation (AMR), 
for scene understanding. 
Drawing on recent advances in AMR parsing, we automatically create datasets of images paired with their AMR graphs, and extend a text-to-AMR parser to the image domain. Additionally, we perform extensive analysis of image AMR graphs, and propose an algorithm for creating image-level AMR graphs from individual caption AMRs. Our analyses and results point to important future research directions for improved scene understanding.
\end{abstract}
\end{comment}

\begin{abstract}
The success of scene graphs for visual scene understanding has brought attention to the benefits of abstracting a visual input (e.g., image) into a structured representation, where entities (people and objects) are nodes connected by edges specifying their relations. Building these representations, however, requires expensive manual annotation in the form of images paired with their scene graphs or frames. These formalisms remain limited in the nature of entities and relations they can capture.
In this paper, we propose to leverage a widely-used meaning representation in the field of natural language processing, the Abstract Meaning Representation (AMR), to address these shortcomings. Compared to scene graphs, which largely emphasize spatial relationships
, our visual AMR graphs are more linguistically informed, with a focus on higher-level semantic concepts extrapolated from visual input. 
Moreover, they allow us to generate meta-AMR graphs to unify information contained in multiple image descriptions under one representation.
Through extensive experimentation and analysis, we demonstrate that we can re-purpose an existing text-to-AMR parser to parse images into AMRs. Our findings point to important future research directions for improved scene understanding.
\end{abstract}

\section{Introduction}
{\let\thefootnote\relax\footnotetext{*Work done during an internship at Samsung AI Centre - Toronto}}
{\let\thefootnote\relax\footnotetext{{\dag}Work done while at Samsung AI Centre - Toronto}}

\begin{figure*}[t!]
    \centering
    \includegraphics[width=\linewidth]{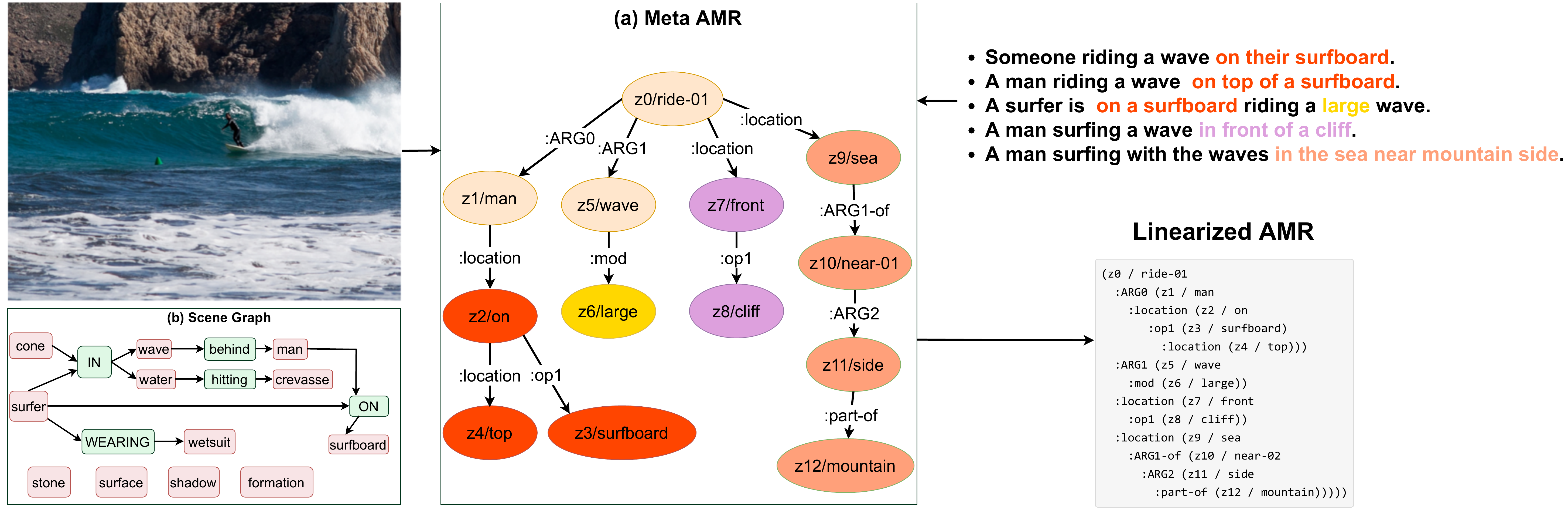}
    \caption{An image\ from MSCOCO and Visual Genome dataset, along with its five human-generated captions, and: (a) an image-level meta-AMR graph capturing its overall semantics, (b) its human-generated scene graph. }
    \label{fig:overview}
\end{figure*}

%% There is debate on what it means to understand an image: briefly mention each of the relevant topics of captioning, SGG, and Situation Recognition (ignore object detection, etc.). Then talk about our proposal. The rest can go to Related Work.

The ability to understand and describe a scene is fundamental for the development of truly intelligent systems, including autonomous vehicles, robots navigating an environment, or even simpler applications such as language-based image retrieval. Much work in computer vision has focused on two key aspects of scene understanding, namely, recognizing entities, including object detection \cite{liu2016ssd,ren2015faster,carion2020end,liu2020deep} and activity recognition \cite{herath2017going,kong2022human,li2018flow,gao2018im2flow}, as well as understanding how entities are related to each other, e.g., human--object interaction \cite{hou2020visual,zou2021end} and relation detection \cite{lu2016visual,zhang2017visual,zellers.etal.2018}. 
%\todo{Vladimir}
%More recently, scene graphs encoding the detected objects and their relationships are widely adopted as a semantic formalism for representing the content of an image \cite{johnson.etal.2015}. Generating scene graphs from an image requires expensive and detailed annotations. As such, others have proposed image captioning as a proxy to image understanding \cite{stefanini.etal.2021.survey}.
%survey paper on image captioning, e.g., https://arxiv.org/abs/2107.06912
%plus original non-DN-based and DN-based image captioning work

A natural way of representing scene entities and their relations is in graph form, so it is perhaps unsurprising that a lot of work has focused on graph-based scene representations and especially on scene graphs \cite{johnson.etal.2015}. Scene graphs encode the salient regions in an image (mainly, objects) as nodes, and the relations among these (mostly spatial in nature) as edges, both labelled via natural language tags; see Fig.~\ref{fig:overview}(b) for an example scene graph.
Along the same lines, \citet{yatskar.etal.2016} propose to represent a scene as a semantic role labelled frame, drawn from FrameNet~\cite{framenet} --- a linguistically-motivated approach that draws on semantic role labelling literature. 

%\citet{yatskar.etal.2016} go a step further and propose situation recognition, or visual semantic role labelling, as a linguistically-motivated approach to image understanding. Given an image, they draw on FrameNet~\cite{framenet} to predict the situation depicted in the image as the combination of a predicate (encoding the main event or activity), and an instantiated frame specifying the participants and their semantic roles. Situation recognition generalizes activity recognition and human--object interaction, focusing on recognizing a {\it single} main activity and detecting its {\it frame-specific} participants (see Fig.~\ref{fig:overview}(c)).
Scene graphs and situation frames can capture important aspects of an image, yet they are limited in important ways. They both require expensive manual annotation in the form of images paired with their corresponding scene graphs or frames. Scene graphs in particular also suffer from being limited in the nature of entities and relations that they capture (see Section~\ref{s:analysis} for a detailed analysis). Ideally, we would like to capture event-level semantics (same as in situation recognition) but as a structured graph that captures a diverse set of relations and goes beyond low-level visual semantics.

%others have proposed image captioning as a proxy to image understanding \cite{stefanini.etal.2021.survey}
Inspired by the linguistically-motivated image understanding research, we propose to represent images using a well-known graph formalism for language understanding, i.e., Abstract Meaning Representations \cite[AMRs][]{banarescu.etal.2013}. Similarly to (visual) semantic role labeling, AMRs also represent ``who did what to whom, where, when, and how?'' \cite{marquez-etal-2008-special}, but in a more structured way via transforming an image into a graph representation.
AMRs not only encode the main events, their participants and arguments, as well as their relations (as in semantic role labelling/situation recognition), but also relations among various other participants and arguments; see Fig.~\ref{fig:overview}(a). 
%We argue that visual semantic parsing into AMRs results in a much richer representation, combining aspects of scene graphs, captions, and situations. Moreover, 
Importantly, AMR is a broadly-adopted and dynamically evolving formalism \cite[e.g.,][]{bonial.etal.2020-dialamr,bonn.etal.2020-spatialamr,naseem.etal.2021}, and AMR parsing is an active and successful area of research \cite[e.g.,][]{zhang.etal.2019-amr,bevilacqua.etal.2021,xia.etal.2021-stackedamr,drozdov.etal.2022-amr}. Finally, given the high quality of existing AMR parsers (for language), we do not need manual AMR annotations for images, and can rely on existing image--caption datasets to create high quality silver data for image-to-AMR parsing. In summary, we make the following contributions:

%Specifically, we propose to adapt AMR \cite{banarescu.etal.2013}, which is a widely-used graph formalism for semantic parsing of language, to the visual domain. 
%AMRs are rooted, directed, acyclic graphs with labels on edges (relations) and leaves (concepts) taken from PropBank \cite{palmer.etal.2005}. Recently, a number of high-performing neural network models have been developed for parsing a sentence into an AMR graph \cite{bevilacqua.etal.2021,xia.etal.2021}.
%Our goal is to devise a visual semantic parser that can learn to represent the semantics of an image using an AMR graph. Two key challenges that we need to address include: (i) adapting a text-to-AMR model to learning AMRs from images, and (ii) achieving this goal in the absence of ground-truth annotations of images with their corresponding AMR graphs.
%\hl{VP: should there be a contribution/challenge related to meta-AMR; typical AMRs are per-sentence but we need meta-AMR here.}
%Summary of our contributions ...
\vspace{-0.3cm}
\begin{itemize}[leftmargin=*]
    \item We introduce the novel problem of parsing images into Abstract Meaning Representations, a widely-adopted linguistically-motivated graph formalism; and propose the first image-to-AMR parser model for the task. \vspace{-0.3cm}
    \item We present a detailed analysis and comparison between scene graphs and AMRs with respect to the nature of entities and relations they capture, results of which further motivates research in the use of AMRs for better image understanding. \vspace{-0.3cm}
    \item Inspired by work on multi-sentence AMR, we propose a graph-to-graph transformation algorithm that combines the meanings of several image caption descriptions into image-level meta-AMR graphs. The motivation behind generating the meta-AMRs is to build a graph that covers most of entities, predicates, and semantic relations contained in the individual caption AMRs.\vspace{-0.3cm}
\end{itemize}

% (i) We introduce the novel problem of parsing images into Abstract Meaning Representations, a widely-adopted linguistically-motivated graph formalism; and propose the first image-to-AMR parser model for the task.
% (ii) We present a detailed analysis and comparison between scene graphs and AMRs with respect to the nature of entities and relations they capture, results of which further motivates research in the use of AMRs for better image understanding.
% (iii) Inspired by work on multi-sentence AMR, we propose a graph-to-graph transformation algorithm that combines the meanings of several image caption descriptions into image-level meta-AMR graphs. The motivation behind generating the meta-AMRs is to build a graph that covers most of entities, predicates, and semantic relations contained in the individual caption AMRs.
%\hl{Mohamed: Do we add a contribution that we will be releasing AMRs and meta-AMRs for MSCOCO?} 
%--- AF: I rather not since we may not get approval from legal.

%(\textcolor{blue}{the meta-AMR follows the same format with AMR format, and is expected to cover most entities, predicates and semantic relations between them within a single sentence meaning representation space}), and \textcolor{blue}{our results} 
%We show that these graphs can in principle be used to generate more descriptive captions. 

Our analyses suggest that AMRs encode aspects of an image content that are not captured by the commonly-used scene graphs. Our initial results on re-purposing a text-to-AMR parser for image-to-AMR parsing, as well as on creating image-level meta-AMRs, point to exciting future research directions for improved scene understanding.
\section{Motivation: AMRs vs. Scene Graphs}\label{s:analysis}

Scene graphs (SGs) are a widely-adopted graph formalism for representing the semantic content of an image. Scene graphs have been shown useful for various downstream tasks, such as image captioning \cite{yang2019auto,li2019know,zhong2020comprehensive}, visual question answering \cite{zhang2019empirical,hildebrandt2020scene,damodaran2021understanding}, and image retrieval \cite{johnson2015image,schuster2015generating,wang2020cross,schroeder2020structured}.
However, learning to automatically generate SGs requires expensive manual annotations (object bounding boxes and their relations). SGs were also shown to be highly biased in the entity and relation types that they capture. For example, an analysis by \citet{zellers.etal.2018} reveals that clothing (e.g., {\it dress}) and object/body parts (e.g., {\it eyes}, {\it wheel}) make up over one-third of entity instances in the SGs corresponding to the Visual Genome images~\cite{krishnavisualgenome}, and that more than $90$\% of all relation instances belong to the two categories of geometric (e.g., {\it behind}) and possessive (e.g., {\it have}).

One advantage of AMR graphs is that we can draw on supervision through captions associated with images. Nonetheless, the question remains as to what types of entities and relations are encoded by AMR graphs, and how these differ from SGs. To answer this question, we follow an approach similar to \citet{zellers.etal.2018}, and categorize entities and relations in SG and AMR graphs corresponding to a sample of ~$50$K images. We use the same categories as \citeauthor{zellers.etal.2018}, but add a few new ones to capture relation types specific to AMRs, namely, Attribute ({\it small}), Quantifier ({\it few}), Event ({\it soccer}), and AMR specific ({\it date-entity}). Details of our categorization process are provided in Appendix~\ref{sec:appendix}.

%Note that, unlike for SGs where nodes represent entities and edges semantic relations, in AMRs, nodes represent both entities and relations (edges in AMRs stand for semantic roles). 

%% Neural Motifs paper does the following: 
%(a) how different types of relations correlate with different objects, and
%(b) how higher order graph structures recur over different scenes. 

%\hl{VP: this section (or the part that describes the annotation details) could go to appendix, if more space is needed.}

\begin{figure}[t!]
    % \centering
    \subfigure{%
      \includegraphics[clip,width=\columnwidth]{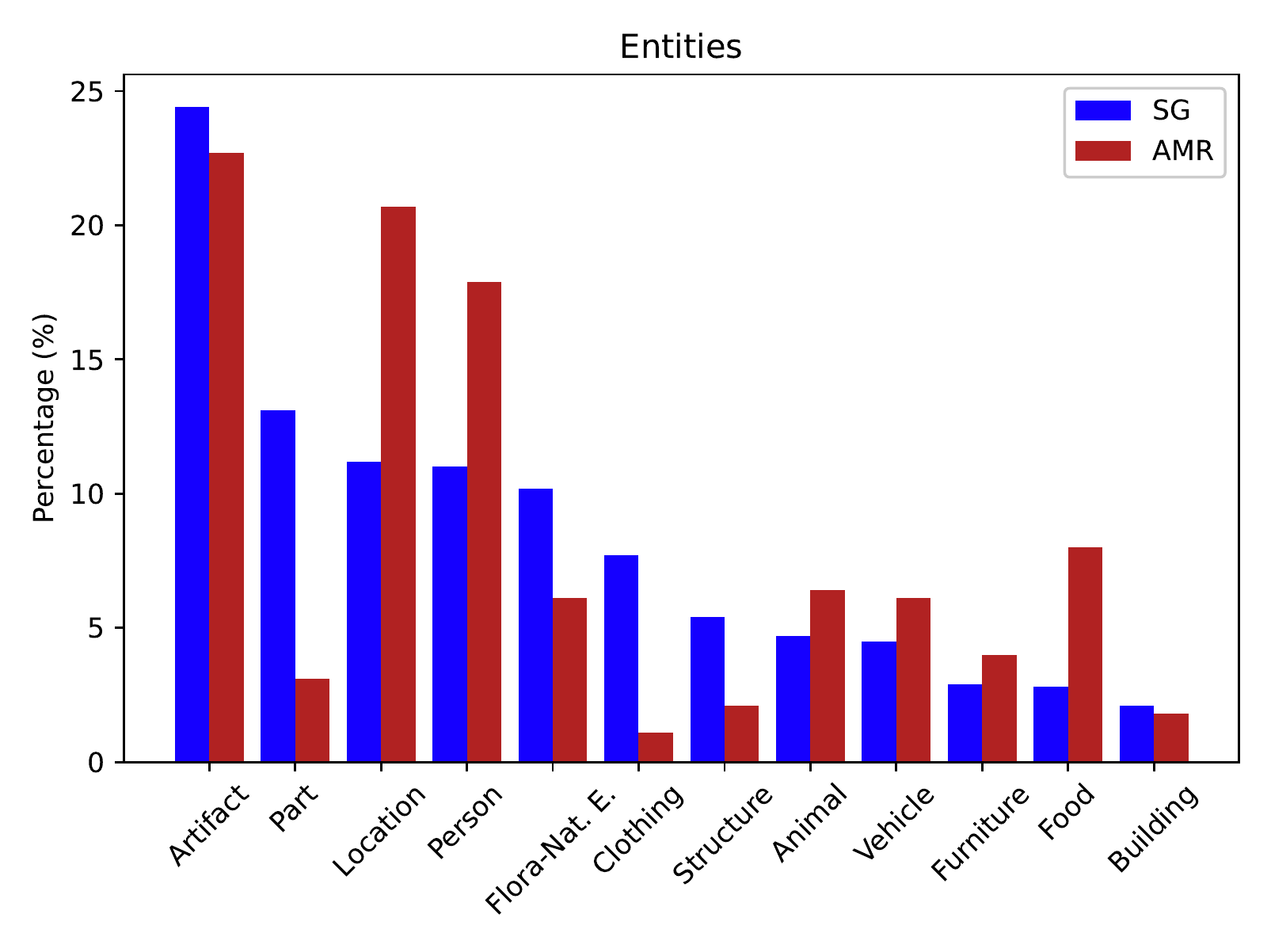}%
    }
    
    \subfigure{%
      \includegraphics[clip,width=\columnwidth]{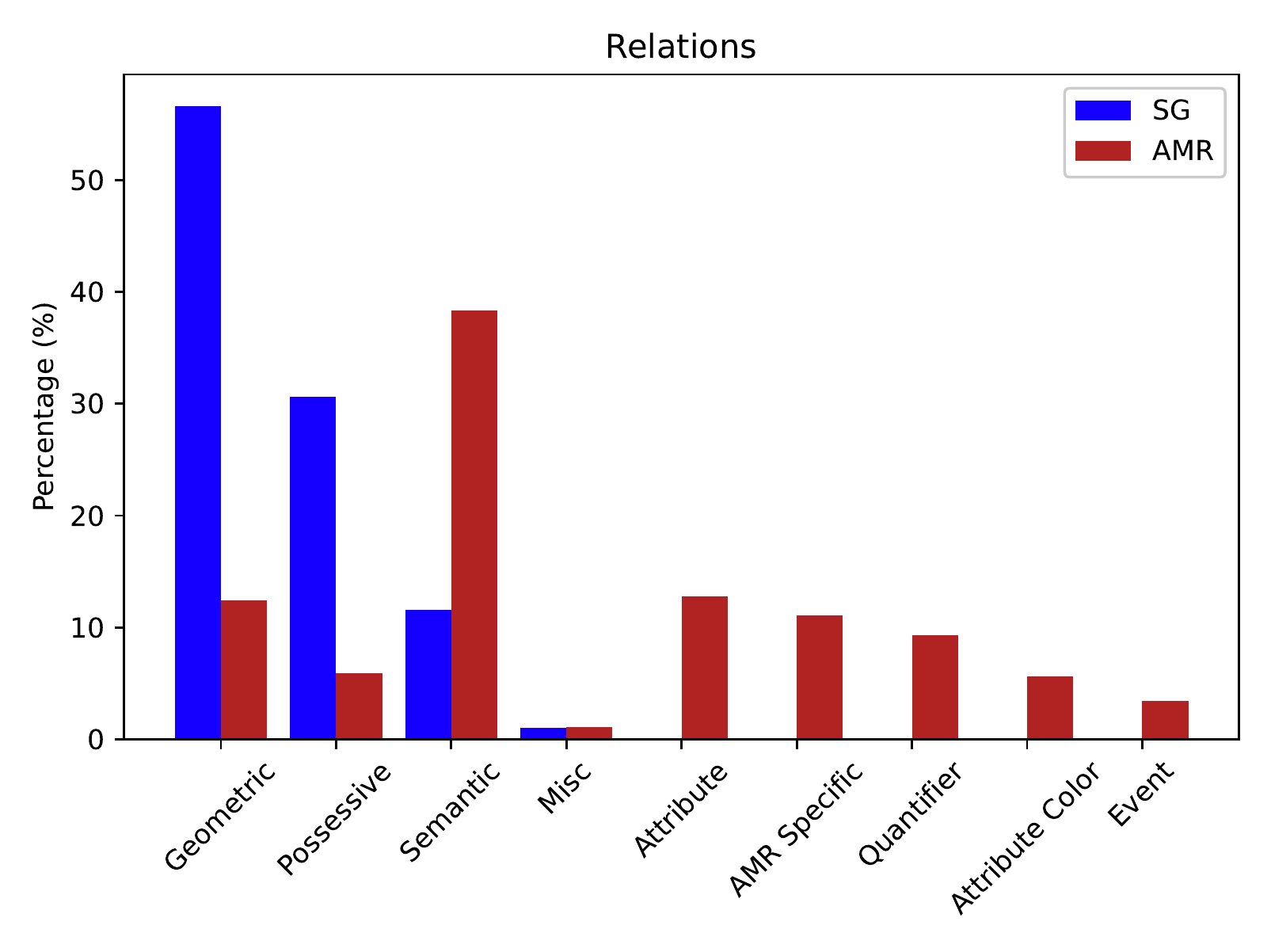}%
    }
    \caption{Statistics on a selected set of top-frequency Entity and Relation categories, extracted from the AMR and SG graphs corresponding to around $50$K images that appear in both Visual Genome and MSCOCO.}
    \label{fig:amr_sg_bar_plots}
\end{figure}

Figure~\ref{fig:amr_sg_bar_plots} shows the distribution of instances for each Entity and Relation category, compared across SG and AMR graphs. 
%\hl{VP: one could even compute a quant score, e.g., entropy, for SG and AMR, to support the claim that one is more diverse than the other.} 
% \todo{Kalliopi}
%{\color{blue}Moreover, if we view the Entity, Relation categories as discrete random variables we can compute their entropy. For the AMR the Entity entropy is $2.13$ and the Relation $1.85$, whereas for the SGs is $2.25$ and $0.99$ accordingly. 
AMRs tend to encode a more diverse set of relations, and in particular capture more of the abstract semantic relations that are missing from SGs. 
%\hl{VP: okay, maybe remove these numbers and focus on qualitative analysis.  Differences in Relations are very pronounced so no need to justify quantitatively. For Entities, could the differences in Location and Person be ascribed to higher level semantics of AMRs? }
This is expected because our caption-generated AMRs by design capture the essential meaning of the image descriptions and, as such, encode how people perceive and describe scenes. 
In contrast, SGs are designed to capture the content of an image, including regions representing objects and (mainly spatial/geometric) visually-observable relations; see Fig.~\ref{fig:overview} for SG and AMR graphs corresponding to an image. In the context of Entities, and a major departure from SGs, (object/body) parts are less frequently encoded in AMRs, pointing to the well-known whole-object bias in how people perceive and describe scenes \cite{markman.1990,fei2007we}. In contrast, location is more frequent in AMRs.
%\hl{VP: this reference is about perception; is there one about verbal or written description?}
 
 The focus of AMRs on abstract content suggests that they have the potential for improving down-stream tasks, especially when the task requires an understanding of the higher level semantics of an image.  Interestingly, a recent study showed that using AMRs as an intermediate representation for textual SG parsing helps improve the quality of the parsed SGs \citep{choi.etal.2022-scene}, even though AMRs and SGs encode qualitatively different information.  Since AMRs tend to capture higher level semantics, we propose to use them as the final image representation. The question remains as to how difficult it is to directly learn such representations from images. The rest of the paper focuses on answering this question.

%\begin{comment}
%\end{comment}
%\section{Related Work}
%\input{relatedwork}
\section{Method}\label{s:method}

\subsection{Parsing Images into AMR Graphs}\label{ss:image2AMR}

\begin{figure*}[t!]
    \centering
    \includegraphics[width=\linewidth]{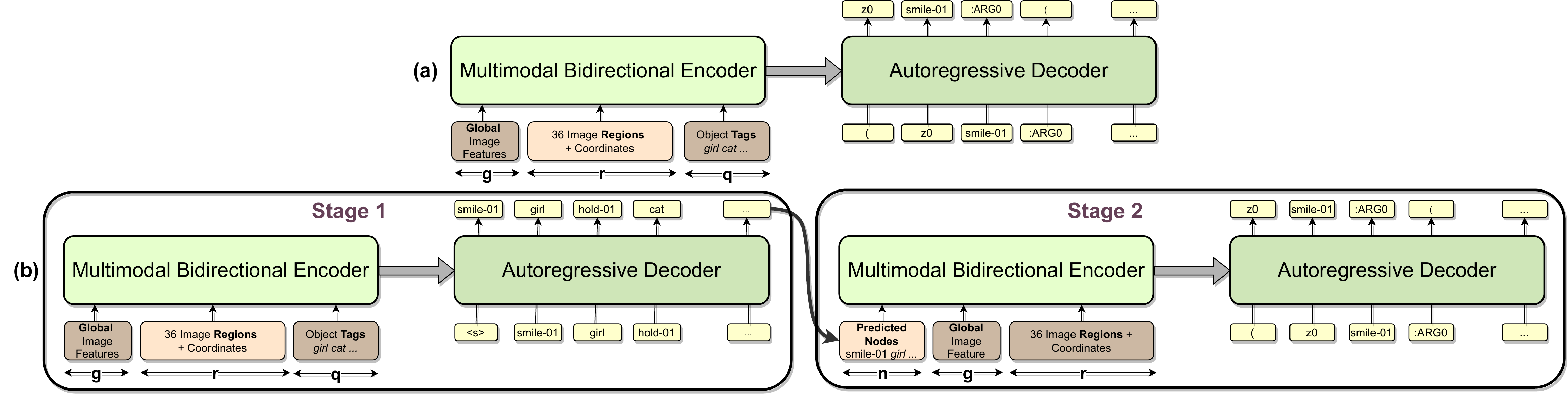}
    \caption{
    Model architecture for our two image-to-AMR models: (a) \textsc{Img2Amr$_\mathrm{direct}$}: A direct model that uses a single seq2seq encoder--decoder to generate linearlized AMRs from input images; and (b) \textsc{Img2Amr}$_\mathrm{2stage}$: A two-stage model containing two independent seq2seq components. $\bf{g}$ and $\bf{r}$ stand for global and region features, $\bf{q}$ for tag embeddings, and $\bf{n}$ for the embeddings of the predicted nodes. The input and output space of the decoders come from the AMR vocabulary.}
    \label{fig:model}
\end{figure*}

We develop image-to-AMR parsers based on a state-of-the-art seq2seq text-to-AMR parser, \textsc{Spring} \cite{bevilacqua.etal.2021}, and a multimodal VL-\textsc{Bart} \cite{cho.etal.2021-vlbart}.
Both are transformer-based architectures with a bi-directional encoder and an auto-regressive decoder.
\textsc{Spring} extends a pre-trained seq2seq model, \textsc{Bart} \cite{lewis.etal.2020-bart}, by fine-tuning it on AMR parsing and generation.
Next, we describe our models, input representation, and training.

\paragraph{Models.}
We build two variants of our image-to-AMR parser, as depicted in Fig.~\ref{fig:model}(a) and (b).
% The specifics of the two setups/architectures
\begin{itemize}
    \item Our first model, which we refer to as \textsc{Img2Amr}$_\mathrm{direct}$,  modifies \textsc{Spring} by replacing \textsc{Bart} with its vision-and-language counterpart, VL-\textsc{Bart} \cite{cho.etal.2021-vlbart}. VL-\textsc{Bart} extends \textsc{Bart} with visual understanding ability through fine-tuning on multiple vision-and-language tasks. With this modification, our model can receive visual features (plus text) as input, and generate linearized AMR graphs.
    \item Our second model, inspired by text-to-graph AMR parsers~\cite[e.g.,][]{zhang.etal.2019-amr,xia.etal.2021-stackedamr}, generates linearized AMRs in two stages by first predicting the nodes, and then the relations. Specifically, we first predict the nodes of the linearized AMR for a given image. These predicted nodes are then fed (along with the image) as input into a second seq2seq model that generates a linearized AMR (effectively adding the relations). We refer to this model as \textsc{Img2Amr}$_\mathrm{2stage}$.
\end{itemize}

%It should be noted that this two-stage model is not end-to-end, as the first stage is trained independently, and then the second stage is trained based on the predictions of the first stage. We refer to this model as \textsc{Img2Amr}$_\mathrm{2stage}$; see Fig.~\ref{fig:model}(b).
%\hl{VP: one should comment on why we chose not to do it end-to-end. Or leave the details of model training for later.}

\paragraph{Input Representation.}
To represent images, we follow VL-\textsc{Bart}, which takes the output of Faster R-CNN~\cite{ren2015faster} (i.e., region features and coordinates for $36$ regions) and projects them onto $d=768$ dimensional vectors via two separate fully-connected layers. Faster R-CNN region features are obtained via training for visual object and attribute classification~\cite{anderson.etal.2018-bottomup} on Visual Genome. The visual input to our model is composed of position-aware embeddings for the $36$ regions, plus a global image-level feature ($\bf{r}$ and $\bf{g}$ in Fig.~\ref{fig:model}).
%\todo{$\bf{g}$ is the result of the same fully connected}
To get the position-aware embeddings for the regions, we add together the projected region and coordinate embeddings. To get the global image feature, we use the output of the final hidden layer in ResNet-101~\cite{he.etal.2016-resnet}, which is passed through the same fully connected layer as the regions to obtain a 768-dimensional vector. % (with dummy coordinates representing the size of the whole image).

%We represent the input image as a $d_u$-dimensional feature vector $u \in \mathbb{R}^{d_u}$ of $K$ object regions obtained from a Faster R-CNN~\cite{ren.etal.2015-fasterrcnn} trained for visual object and attribute classification~\cite{anderson.etal.2018-bottomup} on Visual Genome, where $K = 36$, and $d_u = 2048$. The coordinates and area $z \in \mathbb{R}^5$ of each of these regions are provided as well. 
%
%into equal dimensional vectors that are added into $v \in \mathcal{R}^d$, where $v$ is a position-aware embedding of each region, and $d$ is the dimension of the input to the transformer (i.e., $d = 768$). Inspired by recent work on pre-training models on a diverse set of vision-and-language tasks \cite{li.etal.2020-oscar,cho.etal.2021-vlbart}, we use object tags (assigned by the object detector) as an additional textual input, for which the textual embedding $\bf{q}$ is concatenated to the visual input so that the input to the transformer becomes $[\bf{v}, \bf{q}]$. We also use the output of the final hidden pre-classification layer for a ResNet-101~\cite{he.etal.2016-resnet} pre-trained on ImageNet \cite{deng.etal.2009-imagenet} as a global representation of the image $g' \in \mathcal{R}^{P'}$, concatenated to the object regions $[g', \bf{v}']$. We try ablations with and without these additional inputs $g'$ and $\bf{q}$.

% Training/fine-tuning and Loss (why VL-Bart is a good choice)
\paragraph{Training.}
To benefit from transfer learning, we initialize the encoder and decoder weights of both our models from the pre-trained VL-\textsc{Bart}.
This is a reasonable initialization strategy, given that VL-\textsc{Bart} has been pre-trained on input similar to ours. Moreover, a large number of AMR labels are drawn from the English vocabulary, and thus the pre-training of VL-\textsc{Bart} should also be appropriate for AMR generation.
We fine-tune our models on the task of image-to-AMR generation, using images paired with their automatically-generated AMR graphs. We consider two alternative AMR representations: (a) \textit{caption AMRs}, created directly from captions associated with images (see Section~\ref{s:exp} for details); and (b) image-level \textit{meta-AMRs}, constructed through an algorithm we describe below in Section~\ref{ss:metaAMR}. We perform experiments with either caption or meta-AMRs, where we train and test on the same type of AMRs.
For the various stages of training, we use the cross-entropy loss between the model predictions and the ground-truth labels for each token, where the model predictions are obtained greedily, i.e., choosing the token with the maximum score at each step of the sequence generation.

\subsection{Learning per-Image meta-AMR Graphs}\label{ss:metaAMR}

\begin{algorithm}[h]
\caption{\textbf{\textsc{Meta}-AMR Graph Construction}} \scriptsize
\algrenewcommand\algorithmicindent{1em}
\begin{algorithmic}[1]
\footnotesize
    % \Proc {Roy}{$a,b$} 
    \State {\bf Input}: $k$ human-generated image descriptions $\{c_i\}_{i=1}^{k}$ for a given image $i$; a set of pre-defined AMR relation types $\mathcal{R}$;
    \State {\bf Output}: A meta-AMR graph $g_\mathrm{meta}$;
    \State {\bf Initialize}: 
    Generate AMR graphs $\{g_i\}$ for the $k$ descriptions using a pre-trained AMR semantic parser; 
    Initialize $g_m = (\mathcal{N}, \mathcal{E})$ to be the null graph.
    
    \State
	   $\mathcal{N} = \cup_{i=1}^{k} \mathcal{N}_i$
	\For {$i = 1 \sim  k$}
	    % get nodes, attributes, edges
	    \State
	    %$\mathcal{N}_i=$ getNodes$(g_i)$;
	    $\mathcal{E}_i=$ getEdges$(g_i)$ 
	    %\Comment {\textcolor{blue}{obtain relations from $g_i$, and $\mathbf{N}_i$ is a dictionary of directed edges $r$ from source node $n_s$ to target node $n_t$: \{($n_s$, $n_t$): $r$\}}} 
	    
	    % pickup nodes, attributes and edges to merge into metaAMR
	    %\For {$n$ $\in$ $\mathbf{N}_i$}
	    %    \If{$n \notin \mathbf{N}$}
	    %        \State
	    %        $\mathbf{N}$.add$(n)$
	    %   \EndIf
	   %\EndFor
	   
	   % source=s, target=t, r=relation label
	   
	   \For {$(n_{s}, n_{t}): r$ $\in$ $\mathcal{E}_i$}  
	   \Comment {\textcolor{blue} {($n_s$, $n_t$) is a pair of nodes connected via an edge labeled as $r$}}
	        \If{$(n_{s}, n_{t}) \notin \mathcal{E}.$keys() 
	        \WRP $\land$ $(n_{t}, n_{s}) \notin \mathcal{E}.$keys() 
	        \WRP $\land$ $r \in \mathcal{R}$}
	            \State
	            $\mathcal{E}$.add$(\{(n_{s}, n_{t}): r\})$ \Comment {\textcolor{blue}{Add a new edge when neither $(n_{s}, n_{t})$ nor $(n_{t}, n_{s})$ previously included, and $r$ belongs to a pre-selected set of AMR relation types $\mathcal{R}$}} 
	       \EndIf
	   \EndFor
	   
    \EndFor
    \State
    $\mathcal{G}_{m} = $ weaklyConnectedComponents($g_m$)\Comment{\textcolor{blue}{Get all connected components as $g_\mathrm{meta}$ candidates since it should be a connected graph according to the definition of AMR}}
    \State
    $g_\mathrm{meta} =$ getLargestComponent($\mathcal{G}_m$) \Comment{\textcolor{blue}{Get the candidate with the largest number of nodes as it can cover most entities and predicates in the image}}
    \State
    ${g}_\mathrm{meta} =$ refineNodes(${g}_{meta}$)
    \Comment {\textcolor{blue} {Replace node types by their frequent hypernym if available}} 
    \State 
    \Return $g_{meta}$ 
    
\end{algorithmic}
\label{alg:metaAMR}
\end{algorithm}

Recall that, in order to collect a data set of images paired with their AMR graphs, we rely on image--caption datasets such as MSCOCO. Specifically, we use a pre-trained AMR parser to generate AMR graphs from each caption of an image. Images can be described in many different ways, e.g., each image in MSCOCO comes with five different human-generated captions. 
We hypothesize that these captions collectively represent the content of the image they are describing, and as such propose to also combine the caption AMRs into image-level meta-AMR graphs through a merge and refine process that we explain next.

Prior work has used graph-to-graph transformations for merging sentence-level AMRs into document-level AMRs for abstractive and multi-document summarization~\citep[e.g.,][]{liu.etal.2015,liao.etal.2018,naseem.etal.2021}. Unlike in a summarization task, captions do not form a coherent document, but instead collectively describe an image. Inspired by prior work, we propose our graph-to-graph transformation algorithm that learns a unified meta-AMR graph from caption graphs; see Algorithm~\ref{alg:metaAMR}.
%for creating an image-level meta-AMR graph out of $k$ caption-level AMRs.
Specifically, we first merge the nodes and edges from the original set of $k$ caption-level AMRs, only including a pre-defined set of relation/edge labels. We then select the largest connected component of this merged graph, which we further refine by replacing non-predicate nodes by their more frequent hypernyms, when available. The motivation behind this refinement process is to reduce the complexity of the meta-AMR graphs (in terms of their size), which would potentially improve parsing performance. %In addition, this process results in further merging of concepts that have similar meaning, e.g., if one caption refers to the actor as {\em a woman} and the other as {\em tennis player}, ... is it really the case? need to verify later.
An example of a meta-AMR graph generated from caption AMRs is given in Appendix~\ref{a:meta_amr_example}.

AMR graphs of the MSCOCO training captions contain more than $90$ types of semantic relations and more than $21$K node types, with long-tailed distributions; see Fig.~\ref{fig:edge_type_stats} in Appendix~\ref{a:amr_node_dist}. To refine meta-AMR graphs, we only maintain the top-$20$ most frequent relation types that include core roles, such as \textsc{arg0}, \textsc{arg1}, etc., as well as high-frequency non-core roles, such as mod and location. To further refine the graphs, we replace each non-predicate node (e.g., {\it salmon}) with its most frequent hypernym (e.g., {\it fish}) according to WordNet \cite{wordnet.1998}. 
This results in just about $30\%$ reduction in the number of node types (to $15$K). The average complexity of graphs is also reduced from $19$ nodes and $23$ relations to $16$ and $18$, respectively.

\section{Experimental Setup}\label{s:exp}

\paragraph{Data.}

For our task of AMR generation from images, we use an augmented version of the standard MSCOCO image--caption dataset, which is composed of images paired with their captions, automatically generated caption-level linearized AMR graphs, and an image-level linearized meta-AMR graph.
We use the splits established in previous work \cite{Karpathy2015}, containing $113,287$ training, $5000$ \textsc{val}idation, and $5000$ \textsc{test} images, where each image is associated with five manually-annotated captions. 
Following the cross-modal retrieval work involving MSCOCO \cite[e.g.,][]{lee2018stacked}, we use a subset of the \textsc{val} and \textsc{test} sets, containing $1000$ images each. %\textcolor{blue}{following the cross modal retrieval setting~\cite{lee2018stacked}}.
AMR graphs of the captions are obtained by running the \textsc{Spring} text-to-AMR parser \cite{bevilacqua.etal.2021} that is trained on AMR2.0 dataset.\footnote{\scriptsize{\url{https://catalog.ldc.upenn.edu/LDC2017T10}}} The meta-AMR graph is created from the individual AMRs through our merge and refine process described in Algorithm~\ref{alg:metaAMR} of Section~\ref{s:method}.

\paragraph{Parser implementation details.}

We initialize our \textsc{Img2Amr} models from VL-\textsc{Bart}, which is based on \textsc{Bart}\textsubscript{Base}. \textsc{Bart} uses a sub-word tokenizer with a vocabulary size of $50,265$. Following \textsc{Spring}, we expand the vocabulary to include frequent AMR-specific tokens and symbols (e.g., \textsc{:op}, \textsc{arg1}, {temporal-entity}), resulting in a vocabulary size of $53,587$. The addition of AMR-specific symbols in vocabulary improves efficiency by avoiding extensive sub-token splitting.
%\hl{maybe give some examples of such AMR specific tokens/symbols, we just mention a "subset of them" (-> date entities) in Section 2 - motivation AMR vs SGs --> like "and, or" and their meaning merging different "semantic"-sub sentences in a caption, ordinal-entity, temporal-entity etc}
 The embeddings of these additional tokens are initialized by taking the average of the embeddings of their sub-word constituents. The \textsc{Img2Amr}$_\mathrm{direct}$ models are trained for $60$ epochs, while the \textsc{Img2Amr}$_\mathrm{2stage}$ models are trained for $30$ epochs per stage.
%{\color{red}the next statement is not correct:} \hl{(it should be noted that unlike the typical training for MSCOCO, we have one target/meta-AMR per image and not five targets)}. 
We use a batch size of $10$ with gradients being accumulated for $10$ batches (hence an effective batch size of $100$), the batch size was limited due to the length of the linearized meta-AMRs. The optimizer used is RAdam~\cite{liu.etal.2020-radam}, with a learning rate of $10^{-5}$, and a dropout rate of $0.25$. Each experiment is run on one Nvidia V100-32G GPU.
%\hl{Our code is extending the [\textsc{Spring}] code}, which is based on PyTorch [citation] abd Huggingface Transformers [citation].
Model selection is done based on the best \textsc{SemBleu}-1.
\section{Results}

\begin{table*}[t]
\centering
\begin{tabular}{@{}lc|ccc@{}}
\cmidrule{1-5}
 Model    & Train/Test AMRs & \textsc{Smatch}  & \textsc{SemBleu}-1     & \textsc{SemBleu}-2    \\ 
\cmidrule{1-5} 
 \textsc{Img2Amr}$_\mathrm{direct}$  & meta-AMRs & 37.7 $\pm$ 0.2        & 32.6 $\pm$ 0.8 & 15.2 $\pm$ 0.5          \\ \cmidrule{1-5} 
 \textsc{Img2Amr}$_\mathrm{2stage}$ & meta-AMRs  & 38.6 $\pm$ 0.3 &  30.9 $\pm$ 0.4    & 15.6 $\pm$ 0.3 \\ \cmidrule{1-5} 
  \morecmidrules \cmidrule{1-5} 
\textsc{Img2Amr}$_\mathrm{direct}$ & caption AMRs  & 52.3 $\pm$ 0.4  & 68.6 $\pm$ 0.4 & 38.4 $\pm$ 0.8          \\ \cmidrule{1-5} 
%\textsc{Img2Amr}$_\mathrm{2stage}$ & caption AMRs  & -  & - & - \\ \cmidrule{1-5} 
\end{tabular}
\caption{ \textsc{test} results, averaged over $3$ runs, for our \textsc{Img2Amr} models that follow the best setting, when trained and tested on either meta-AMRs or caption AMRs.}
\label{res:img2AMR}
\end{table*}

\begin{table*}[h]
\centering
\begin{tabular}{@{}l|cccc@{}}
\toprule
 Model & \textsc{Bleu}-4 & \textsc{Cide}r & \textsc{Meteor} & \textsc{Spice} \\ \midrule
\textsc{Img2Amr}$_\mathrm{direct}$ + \textsc{Amr2Txt} & 31.7   & 111.7 & 26.8   & 20.4 \\
VL-\textsc{Bart}$^{*}$    & 35.1   & 116.6 & 28.7   & 21.5  \\ 
\bottomrule
\end{tabular}
\caption{Image captioning results on \textsc{test} set, compared with the best reported captioning results for VL-\textsc{Bart}.}
\label{res:amr2caption}
\end{table*}

\subsection{Image-to-AMR Parsing Performance}

We use the standard measures of \textsc{Smatch}~\cite{cai.knight.2013-smatch} and \textsc{SemBleu}~\cite{song.gildea.2019-sembleu} to evaluate our various image-to-AMR models. \textsc{Smatch} compares two AMR graphs by calculating the F1-score between the nodes and edges of these two graphs. This score is calculated after applying a one-to-one mapping of the two AMRs based on their nodes. This mapping is chosen so that it maximizes the F1-score between the two graphs. However, since finding the best exact mapping is NP-complete, a greedy hill-climbing algorithm with multiple random initializations is used to obtain this best mapping. \textsc{SemBleu} extends the \textsc{Bleu}~\cite{papineni2002bleu} metric to AMR graphs, where each AMR node is considered a unigram (used in \textsc{SemBleu-1}), and each pair of connected nodes along with their connecting edge is considered a bigram (used in \textsc{SemBleu-2}). These metrics are calculated between the model predictions and the noisy AMR ground-truth.

We report results on generating caption AMRs (when the models are trained and tested on these AMRs), as well as meta-AMRs.
When evaluating on caption AMR generation, we compare the model output to the five reference AMRs, and report the maximum of these five scores. The intuition is to compare the predicted AMR to the most similar AMR from the five references.
Table~\ref{res:img2AMR} (top two rows) shows the performance of the models on the task of generating meta-AMRs from \textsc{test} images. We perform ablations of the model input combinations on \textsc{val} set (see Section~\ref{ss:ablations} below), and report \textsc{test} results for the best setting, which uses all the input features for both models.
The $\mathrm{2stage}$ model does slightly better on this task, when looking at the \textsc{Smatch} and \textsc{SemBleu}-2 metrics that take the structure of AMRs into account. Note that \textsc{SemBlue}-1 only compares the nodes of the predicted and ground-truth graphs.
%The settings vary the input to the encoders, where in addition to the global image features, we may use region embeddings, tag embeddings (for the first encoder), and node embeddings (for the second encoder of \textsc{Img2Amr}$_\mathrm{2stage}$). 
%For all models, we found that using all input features results in best performance.

Meta-AMR graphs tend to, on average, be longer than individual caption AMRs (${\sim}34$ vs ${\sim}12$ nodes and relations). 
We thus expect the generation of meta-AMRs to be harder than that of caption AMRs. %\hl{Mohamed: is it clear that the sequence length is much larger than 13? (for the edges, brackets, etc)}\cite{}.\hl{add citation} 
Moreover, although we hypothesize that meta-AMRs capture a holistic meaning for an image, the caption AMRs still capture some (possibly salient) aspect of an image content, and as such are useful to predict, especially if they can be generated with higher accuracy. We thus report the performance of our $\mathrm{direct}$ model on generating caption AMRs (when trained on caption AMR graphs); see the final row of Table~\ref{res:img2AMR}. We can see that, as expected, performance is much higher on generating caption AMRs vs. meta-AMRs.

Given that AMRs and natural language are by design closer in the semantic space, unlike for AMRs and images, it is not unexpected that the results for our image-to-AMR task are not comparable with those of SoTA text-to-AMR parsers, including \textsc{Spring}. Our results highlight the challenges similar to those of general image-to-graph parsing techniques, including visual scene graph generation~\cite{zhu.etal.2022.survey}, where there still exists a large gap in predictive model performance.
%\hl{@Vladimir: add something about image to SG, and how that helps interpret our results.}
%These results, although promising, suggest that generating AMRs from images needs further improvement. To better understand the quality of these AMRs, we perform an extrinsic evaluation (using AMRs for generating captions) and report results in the following section.
%\hl{VP: it may be good to state why we deem these results insufficiently good; compared to what?}\todo{compared to text-2-AMR parsers}

%\hl{AF: why only 1-Stage? need to say something}
%Table~\ref{res:indAMR} shows the results on {\color{red} Test} images.
%\hl{AF: we may want to merge Tables 1 and 2, with a double-line in between}

%\begin{table}[htb!]
%\centering
%\begin{tabular}{@{}l|ccc@{}}
%\cmidrule{1-4}
%Model                                   & Smatch        & %\textsc{SemBleu}-1     & \textsc{SemBleu}-2    \\ \cmidrule{1-4} 
%1-Stage   & 53.2          & 70.0 & 39.4          \\ 
%\cmidrule{1-4} 
%Captions -> AMRs  & - & -    & - \\ 
%\cmidrule{1-4} 
%\end{tabular}
%\caption{\hl{val set, to be changed} Test results for image-to-AMR %generation when trained and tested on caption-level AMRs.}
%\label{res:indAMR}
%\end{table}

\subsection{Image-to-AMR for Caption Generation}\label{ss:captioning}

To better understand the quality of our generated AMRs, we use them to automatically generate sentences from caption AMRs (using an existing AMR-to-text model), and evaluate the quality of these generated sentences against the reference captions of their corresponding images. Specifically, we use the \textsc{Spring} AMR-to-text model that we train from scratch on a dataset composed of AMR2.0, plus the training MSCOCO captions paired with their (automatically-generated) AMRs.
% \hl{bit confused}
%Do we use two different AMR2Text models? So one we train, but how did we get the automatically generated captions for training? Did we use a pretrained AMR2Text? If yes, why did we have to re-train a model on the "synthetic/automatic" data?} 
We evaluate the quality of our AMR-generated captions using standard metrics commonly used in the image captioning community, i.e., \textsc{Cide}r~\cite{vedantam2015cider}, \textsc{Meteor}~\cite{denkowski2014meteor}, \textsc{Bleu}-4~\cite{papineni2002bleu}, and \textsc{Spice}~\cite{anderson2016spice}, and compare against VL-\textsc{Bart}'s best captioning performance as reported in the original paper~\cite{cho.etal.2021-vlbart}. Reported in Table~\ref{res:amr2caption}, the results clearly show that the quality of the generated AMRs are such that reasonably good captions can be generated from them, suggesting that AMRs can be used as intermediate representations for such downstream tasks. Future work will need to explore the possibility of further adapting the AMR formalism to the visual domain, as well as the possibility of enriching image AMRs via incorporating additional linguistic or commonsense knowledge, that could potentially result in better quality captions.

\begin{comment}
 These results are not surprising, and point to the inherent difficulty in translating between the visual and the linguistic domains. Indeed, our analysis of the AMR concepts and relations suggest that many of the AMR concepts (that come from captions) do not have a visual counterpart, e.g., quantifiers ({\it some}, {\it many}), and abstract attributes ({\it young}, {\it small}).
%\hl{VP: comment on the "error", or the lack of, that AMR2Text may be introducing?}\todo{Afsaneh}
\end{comment}

\subsection{Performance per Concept Category}

\begin{figure}[t]
    % \centering
    \subfigure{%
      \includegraphics[clip,width=\columnwidth]{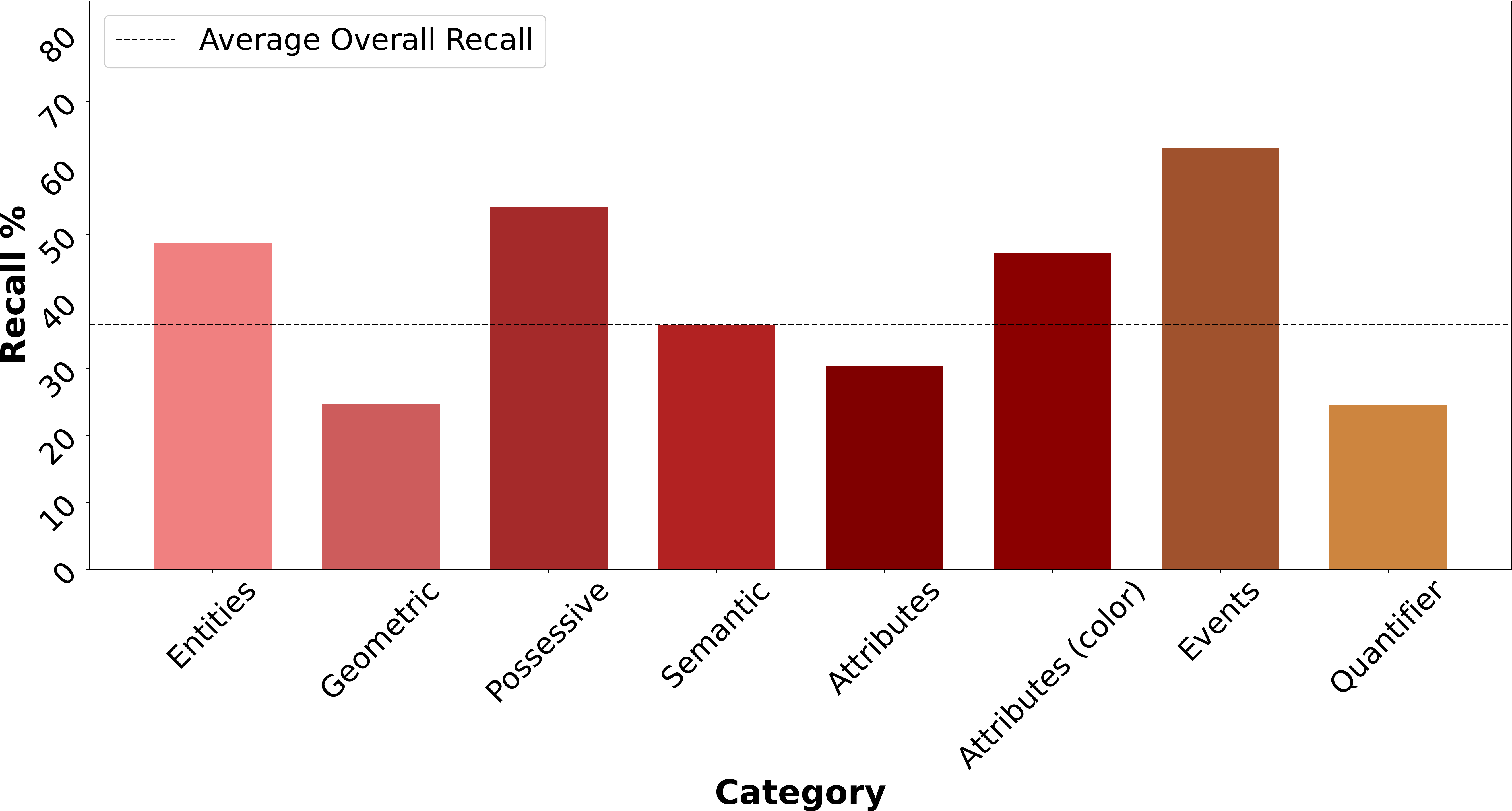}%
    }
    \subfigure{%
      \includegraphics[clip,width=\columnwidth]{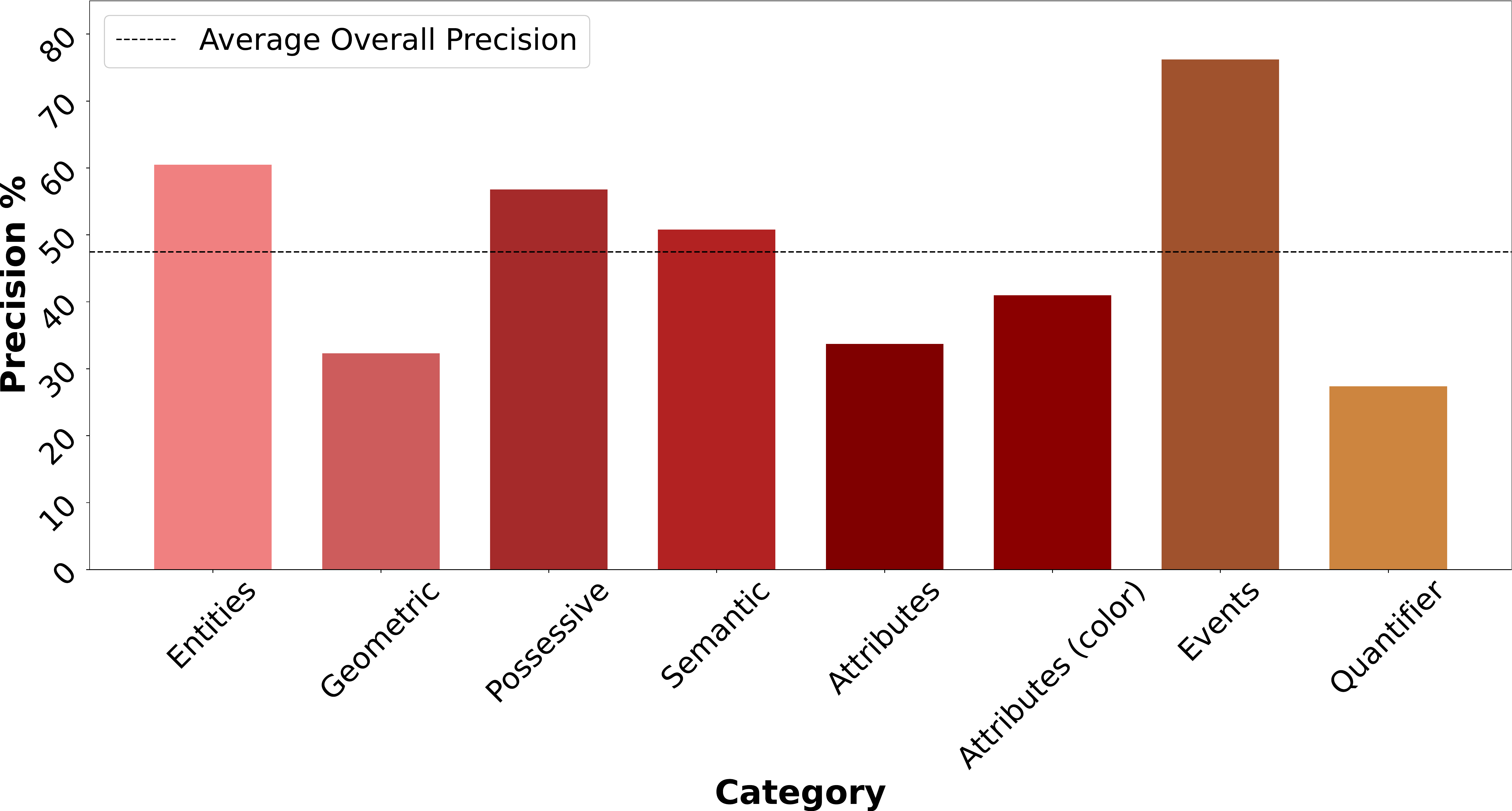}%
    }
    \caption{Node prediction performance on \textsc{val}, for the two-stage model, broken down by category.}
    %\hl{VP: I would keep the same order of labels as that in Fig.2.}}
    \label{fig:nodes_categories_analysis}
\end{figure}

The analysis presented in Section~\ref{s:analysis} suggests many concepts in AMR graphs tend to be on the more abstract (less perceptual) side. We thus ask the following question: What are some of the categories that are harder to predict?
%\todo{what are some of the categories that are harder to predict?} 
%Are some of the entity or relation categories easier to predict than others? 
To answer this question, we look into the node prediction performance of our two-stage model for the different entity and relation categories of Section~\ref{s:analysis}.
Note that this categorization is available for a subset of nodes only. To get the per-category recall and precision values, we take the node predictions of the first stage of the \textsc{Img2Amr}$_\mathrm{2stage}$ model (trained to predict meta-AMR nodes) on the \textsc{val} set. For each \textsc{val} image $i$, we have a set of predicted nodes, which we compare to the set of nodes in the ground-truth meta-AMR associated with the image. When calculating per-category recall/precision values, we only consider nodes that belong to that category. We calculate per-image true positive, true negative, and false positive counts, which are used to obtain the recall and precision using micro-averaging.
\begin{comment}
\begin{equation}
    \mathrm{Recall} = \frac{\sum_{i \in \mathcal{D}} \mathrm{TP}_i}{\sum_i^{N} \mathrm{TP}_i + \sum_i^{N} \mathrm{FN}_i}
\end{equation}
\begin{equation}
    \mathrm{Precision} = \frac{\sum_i^{N} \mathrm{TP}_i}{\sum_i^{N} \mathrm{TP}_i + \sum_i^{N} \mathrm{FP}_i}
\end{equation}
\end{comment}
Fig.~\ref{fig:nodes_categories_analysis} presents the per-category (as well as overall) recall and precision values over the \textsc{val} set.

Interestingly, events (e.g., {\it festival}, {\it baseball}, {\it tennis}) have the highest precision and recall. These are abstract concepts that are largely absent from SGs, suggesting that relying on a linguistically-motivated formalism is beneficial in capturing such abstract aspects of an image content. The event category contains $14$ different types, many referring to sports that have a very distinctive setup, e.g., people wearing specific clothes, holding specific objects, etc. 
%Our models can learn to generate such abstract concepts with high precision and recall, because they appear in the training AMRs that are generated from human-written descriptions likely to mention the event names.
The possibility of encoding such abstract concepts in the training AMRs (generated from human-written descriptions likely to mention the event) helps the model learn to generate them for the relevant images during inference.
%\hl{VP: would be great if one could demonstrate this in select examples / qualitative samples of generated AMRs!} Great suggestion. We will try to find one such example for the camera-ready. Tried to massage the sentence a bit.
The next group with high precision and recall are entities (which are likely to be more closely tied to the image regions), and possessives (containing a small number of high-frequency relations, e.g., {\it have} and {\it wear}). Semantic relations have a decent performance, but contain a diverse number of types, and need to be further analyzed to disentangle the effect of category vs. frequency.

Quantifiers (many of which are related to counting), geometric relations, and attributes seem to be particularly hard to predict. Counting is known to be hard for deep learning models. Geometric relations are much less frequent in AMRs, compared to SGs. Perhaps, we do need to rely on special features (e.g., relative position of bounding boxes) to improve performance on these relations.
%
%It is surprising to see that geometric (spatial) relations are also difficult for our model to predict. \hl{VP: is it?  There are fewer geometric relationships in AMRs than in SGs, so perhaps insufficient data to learn accurate  models.} We may need to rely on special features (relative position of bounding boxes) as well as grounding of entities to alleviate this issue. 
Attributes (such as {\em young}, {\em old}, {\em small}) require the model to learn subtle visual cues.
%Predicting semantic relations (such as {\it eating} or {\it sitting}) may require an overall understanding of the image, and some level of reasoning over the full set of entities and relations. 
In addition to understanding what input features may help improve performance on these categories, we need to further adapt the AMR formalism to the visual domain.

%\hl{VP: are we being too pessimistic about img2AMR performance?  It should be placed in the context of img2SG performance, of the same categories.}\todo{What is SoTA in img2SG?}

%\hl{For possessive we might can explain the better performance since it refers to 5 keywords - wearing, have, include, contain -. Also the events are limited 15 different sub-categories and relate to sports which have a very distinctive setup, field, people wearing specific clothes, holding specific objects etc for which we have the object tags and object bounding boxes as model inputs. Also the fact that the input is focusing on entities explains can explain the very good performance on "Entities" class.}

\begin{comment}
\begin{figure*}[t]
\centering
\begin{subfigure}[b]{0.48\linewidth}
\centering
    \includegraphics[width=\linewidth]{figures/nodes_recall.pdf}
\end{subfigure}
\begin{subfigure}[b]{0.48\linewidth}
\centering
    \includegraphics[width=\linewidth]{figures/nodes_precision.pdf}
\end{subfigure}
\caption{\hl{val set} Node prediction performance for the 2-Stage model for the different categories of nodes.}
    \label{fig:nodes_categories_analysis}
\end{figure*}
\end{comment}

% \begin{figure*}
%     \centering
%     \includegraphics[width=\linewidth]{figures/recall_analysis.png}
%     \caption{Node prediction performance for the 2-Stage model under various conditions, applied to per-image meta-AMR graphs.}
%     \label{tab:recall_ablations}
% \end{figure*}

% \begin{table*}[h!]
%     \centering
%     \begin{tabular}{c|c|c|c|c|c|c}
%     Regions & Tags & Global Vector & Overall Recall & Predicates - have \& wear \& colors & COCO classes & Non-Predicates – COCO Classes \\
%         \hline
%         No & Yes & No & 30.4 & 42.8 & 29.7 & 4.5 \\
%         \hline
%         Yes & No & No & 34.5 & 47.1 & 33.1 & 5.5 \\
%         \hline
%         No & No & Yes & - & - & - & - \\
%         \hline
%         Yes & Yes & No & 35.8 & 49.0 & 34.3 & 6.0  \\
%         \hline
%         Yes & No & Yes & 35.2 & 47.5 & 33.9 & 5.9 \\
%         \hline
%         No & Yes & Yes & - & - & - & - \\
%         \hline
%         Yes & Yes & Yes & 36.7 & 48.4 & 35.6 & 5.9 \\

%     \end{tabular}
%     \caption{Merged AMRs (V2) recall analysis of the nodes prediction (first stage of the 2\_stage\_model)}
%     \label{tab:recall_ablations}
% \end{table*}

\begin{figure*}[!ht]
%    \subfigure[A group of people in the dirt standing on top of a dirt beach with a few flying kites in it.]{%
%      \includegraphics[clip, width=0.22\textwidth, height=0.24\textwidth]{figures/qualitative/1.jpeg}
%    }
    \subfigure[A couple of giraffe standing next to each other in a field near rocks walking in grass in a grassy area.]{%
      \includegraphics[clip, width=0.225\textwidth, height=0.24\textwidth]{figures/qualitative/2.jpeg}
    }
    \subfigure[A yellow and blue fire hydrant on a city street in front at an intersection sitting on the side of the road near a traffic position.]{%
      \includegraphics[clip, width=0.225\textwidth, height=0.24\textwidth]{figures/qualitative/3.jpeg}
    }
    \subfigure[A large long passenger train going across a wooden beach plate, traveling and passing by water.]{%
      \includegraphics[clip, width=0.225\textwidth, height=0.24\textwidth]{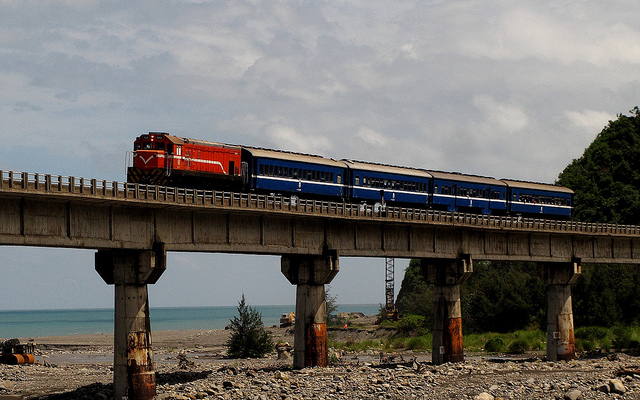}
    }
    \subfigure[A woman sitting at a table eating a sandwich and holding a hot dog in a building smiling while eating.]{%
      \includegraphics[clip, width=0.225\textwidth, height=0.24\textwidth]{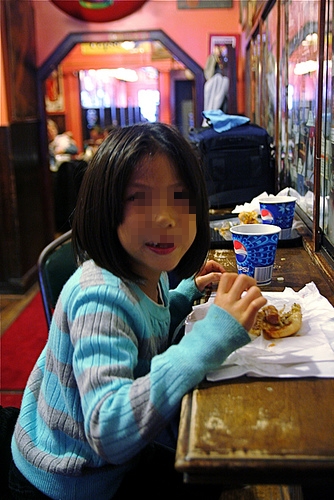}
    }
%    \subfigure[A herd of zebra standing next to each other on a dry grass field grazing on top of a ground field.]{%
%      \includegraphics[clip, width=0.22\textwidth, height=0.24\textwidth]{figures/qualitative/5.jpeg}
%    }
    \subfigure[A white area filled with lots of different kinds of donuts with various toppings sitting on them.]{%
      \includegraphics[clip, width=0.225\textwidth, height=0.24\textwidth]{figures/qualitative/7.jpeg}
    }
    \hspace{0.015\textwidth}
    \subfigure[A group of people sitting around at a dining table with water posing for a picture.]{%
      \includegraphics[clip, width=0.225\textwidth, height=0.24\textwidth]{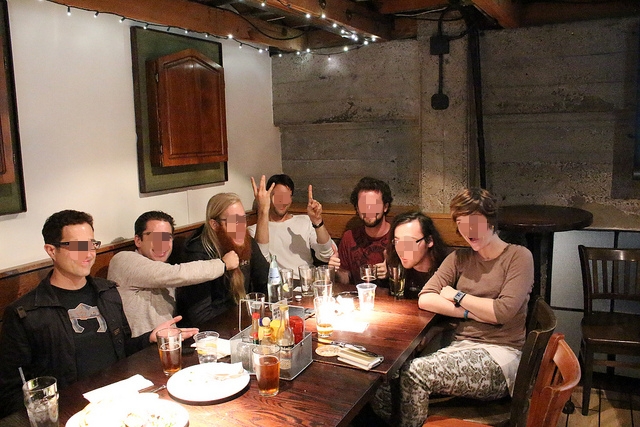}
    }
%    \subfigure[Driving a blue and white person deck car traveling down a street with an advertisement on the side.]{%
%      \includegraphics[clip, width=0.22\textwidth, height=0.24\textwidth]{figures/qualitative/9.jpeg}
%   }
    \hspace{0.015\textwidth}
    \subfigure[A person in a red jacket cross country skiing down a snow covered ski slope with a couple of people riding skis and walking on the side of the snowy mountain.]{%
      \includegraphics[clip, width=0.225\textwidth, height=0.24\textwidth]{figures/qualitative/10.jpeg}
    }
    \hspace{0.015\textwidth}
    \subfigure[A person in black shirt sitting at a table in a building with a plate of food with and smiling while having meal.]{%
      \includegraphics[clip, width=0.225\textwidth, height=0.24\textwidth]{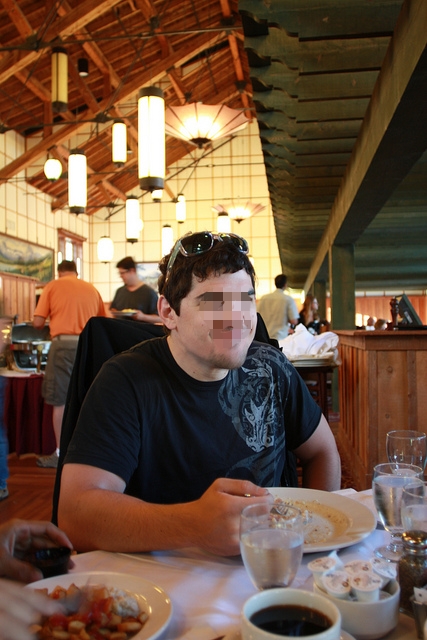}
    }
    \caption{A sample of images, along with descriptive captions automatically generated from the meta-AMRs predicted by our \textsc{Img2Amr}$_\mathrm{direct}$ model. Refer to Section~\ref{ss:qualitative_amrs} for the generated meta-AMRs. The url and license information for each of these images is available in  Section~\ref{ss:qualitative_amrs}. Faces were blurred for privacy.}
    \label{fig:qualitative}
\end{figure*}

\subsection{Qualitative Samples: Generating Descriptive Captions from meta-AMRs}\label{ss:qualitative}

In Section~\ref{ss:captioning}, we showed that caption AMRs produced by our \textsc{Img2Amr} model can be used to generate reasonably good quality captions via an AMR-to-text model. Here, we provide samples of how meta-AMRs can be used as rich intermediate representations for generating descriptive captions; see Fig.~\ref{fig:qualitative} and Section~\ref{ss:qualitative_amrs}. To get these captions, we apply the same AMR-to-text model that we trained as described in Section~\ref{ss:captioning} to the meta-AMRs predicted by our \textsc{Img2Amr}$_\mathrm{direct}$ model. Captions generated from meta-AMRs tend to be longer than the original human-generated captions, and contain much more details about the scene. These captions, however, sometimes contain repetitions of the same underlying concept/relation (though using different wordings), e.g., caption (a) contains both {\it in grass} and {\it in a grassy area}.
We also see that our hypernym replacement sometimes results in using a more general term in place of a more specific but more appropriate term, e.g., {\it woman} instead of {\it girl} in (d). Nonetheless, these results generally point to the usefulness of AMRs and especially meta-AMRs for scene representation and caption generation.
\section{Discussion and Outlook}

%\paragraph{Adapting AMRs to the Visual Domain.}

%Refining AMRs, plus improvements to the meta-AMR generation. 
%(images on the web also often have several descriptions, so this meta-AMR creation is not specific to datasets such as MSCOCO.)

%Can we draw on SGs to ground AMR concepts? What would the advantage of such ground be?

In this paper, we proposed to use a well-known linguistic semantic formalism, i.e., Abstract Meaning Representation (AMR) for scene understanding. We showed through extensive analysis the advantages of AMR vs. the commonly-used visual scene graphs, and proposed to re-purpose existing text-to-AMR parsers for image-to-AMR parsing. Additionally we proposed a graph transformation algorithm that merges several caption-level AMR graphs into a more descriptive meta-AMR graph. Our quantitative (intrinsic and extrinsic) and qualitative evaluations demonstrate the usefulness of (meta-)AMRs as a scene representation formalism.

Our findings point to a few exciting future research directions. Our image-to-AMR parsers can be improved by incorporating richer visual features, a better understanding of the entity and relation categories that are particularly hard to predict for our current models, as well as drawing on methods used for scene graph generation~\cite[e.g.,][]{zellers.etal.2018,zhu.etal.2022.survey}. Our meta-AMR generation algorithm can be further tuned to capture visually-salient information (e.g., quantifiers are too hard to learn from images, and perhaps can be dropped from a visual AMR formalism).

Our qualitative samples of captions generated from meta-AMRs show their potential for generating descriptive and/or controlled captions.
Controllable image captioning has received a great deal of attention lately
\cite[e.g.,][]{cornia.etal.2019,chen.etal.2020,chen.etal.2021}. It focuses on the use of subjective control, including personalization and style-focused caption generation, as well as objective control on content (controlling what the caption is about, e.g., focused on a set of regions), or on the structure of the output sentence (e.g., controlling sentence length). We believe that by using AMRs as intermediate scene representations, we can bring together the work on these various types of control, as well as draw on the literature on controllable natural language generation~\cite{zhang.etal.2022-survey} for advancing research on rich caption generation.

%\paragraph{Application in Controllable Captioning.}

%Adding common sense information (from knowledge bases, such as ATOMIC and ConceptNet)
%Video2Common Sense: \cite{fang.etal.2020}

%Controllable and grounded captioning: \cite{cornia.etal.2019}
%Controlled captioning with abstract scene graphs \cite{chen.etal.2020}

%human-like controllable captioning with semantic roles \cite{chen.etal.2021}
% subjective control focuses on personalization and style of captions generated, whereas objective control focuses on either content (that controls what the caption is about, e.g., set of regions, SGs, visual relations), or structure of output sentence (controlling the length of the caption, part-of-speech of the sequence, etc.)
% chen.etal.2021 proposes to use content-based control signals, but in such a way that the regions to control the caption generation are compatible for a given verb (and its semantic roles).

%Controllable text generation in general \url{https://arxiv.org/abs/2201.05337}

%\paragraph{Extending to the Video Domain.}

%Require addition of tracking and temporal relations, etc.

% \clearpage`
\bibliography{anthology,custom}

\begin{thebibliography}{61}
\expandafter\ifx\csname natexlab\endcsname\relax\def\natexlab#1{#1}\fi

\bibitem[{Anderson et~al.(2016)Anderson, Fernando, Johnson, and
  Gould}]{anderson2016spice}
Peter Anderson, Basura Fernando, Mark Johnson, and Stephen Gould. 2016.
\newblock Spice: Semantic propositional image caption evaluation.
\newblock In \emph{European Conference on Computer Vision}.

\bibitem[{Anderson et~al.(2018)Anderson, He, Buehler, Teney, Johnson, Gould,
  and Zhang}]{anderson.etal.2018-bottomup}
Peter Anderson, Xiaodong He, Chris Buehler, Damien Teney, Mark Johnson, Stephen
  Gould, and Lei Zhang. 2018.
\newblock Bottom-up and top-down attention for image captioning and visual
  question answering.
\newblock In \emph{IEEE Conference on Computer Vision and Pattern Recognition}.

\bibitem[{Banarescu et~al.(2013)Banarescu, Bonial, Cai, Georgescu, Griffitt,
  Hermjakob, Knight, Koehn, Palmer, and Schneider}]{banarescu.etal.2013}
Laura Banarescu, Claire Bonial, Shu Cai, Madalina Georgescu, Kira Griffitt, Ulf
  Hermjakob, Kevin Knight, Philipp Koehn, Martha Palmer, and Nathan Schneider.
  2013.
\newblock Abstract meaning representation for sembanking.
\newblock In \emph{7th Linguistic Annotation Workshop and Interoperability with
  Discourse}.

\bibitem[{Bevilacqua et~al.(2021)Bevilacqua, Blloshmi, and
  Navigli}]{bevilacqua.etal.2021}
Michele Bevilacqua, Rexhina Blloshmi, and Roberto Navigli. 2021.
\newblock One {SPRING} to rule them both: Symmetric {AMR} semantic parsing and
  generation without a complex pipeline.
\newblock In \emph{Association for the Advancement of Artificial Intelligence}.

\bibitem[{Bonial et~al.(2020)Bonial, Donatelli, Abrams, Lukin, Tratz, Marge,
  Artstein, Traum, and Voss}]{bonial.etal.2020-dialamr}
Claire Bonial, Lucia Donatelli, Mitchell Abrams, Stephanie~M. Lukin, Stephen
  Tratz, Matthew Marge, Ron Artstein, David Traum, and Clare Voss. 2020.
\newblock Dialogue-{AMR}: {A}bstract {M}eaning {R}epresentation for dialogue.
\newblock In \emph{Proceedings of the 12th Language Resources and Evaluation
  Conference}.

\bibitem[{Bonn et~al.(2020)Bonn, Palmer, Cai, and
  Wright-Bettner}]{bonn.etal.2020-spatialamr}
Julia Bonn, Martha Palmer, Zheng Cai, and Kristin Wright-Bettner. 2020.
\newblock Spatial {AMR}: Expanded spatial annotation in the context of a
  grounded {M}inecraft corpus.
\newblock In \emph{Proceedings of the 12th Language Resources and Evaluation
  Conference (LREC)}.

\bibitem[{Cai and Knight(2013)}]{cai.knight.2013-smatch}
Shu Cai and Kevin Knight. 2013.
\newblock {S}match: an evaluation metric for semantic feature structures.
\newblock In \emph{51st Annual Meeting of the Association for Computational
  Linguistics}.

\bibitem[{Carion et~al.(2020)Carion, Massa, Synnaeve, Usunier, Kirillov, and
  Zagoruyko}]{carion2020end}
Nicolas Carion, Francisco Massa, Gabriel Synnaeve, Nicolas Usunier, Alexander
  Kirillov, and Sergey Zagoruyko. 2020.
\newblock End-to-end object detection with transformers.
\newblock In \emph{European Conference on Computer Vision}.

\bibitem[{Chen et~al.(2021)Chen, Jiang, Xiao, and Liu}]{chen.etal.2021}
Long Chen, Zhihong Jiang, Jun Xiao, and Wei Liu. 2021.
\newblock Human-like controllable image captioning with verb-specific semantic
  roles.
\newblock In \emph{Proceedings of the IEEE Conference on Computer Vision and
  Pattern Recognition}.

\bibitem[{Chen et~al.(2020)Chen, Jin, Wang, and Wu}]{chen.etal.2020}
Shizhe Chen, Qin Jin, Peng Wang, and Qi~Wu. 2020.
\newblock Say as you wish: Fine-grained control of image caption generation
  with abstract scene graphs.
\newblock In \emph{Proceedings of the IEEE Conference on Computer Vision and
  Pattern Recognition}.

\bibitem[{Cho et~al.(2021)Cho, Lei, Tan, and Bansal}]{cho.etal.2021-vlbart}
Jaemin Cho, Jie Lei, Hao Tan, and Mohit Bansal. 2021.
\newblock Unifying vision-and-language tasks via text generation.
\newblock In \emph{International Conference on Machine Learning}.

\bibitem[{Choi et~al.(2022)Choi, Heo, Punithan, and
  Zhang}]{choi.etal.2022-scene}
Woo~Suk Choi, Yu-Jung Heo, Dharani Punithan, and Byoung-Tak Zhang. 2022.
\newblock Scene graph parsing via {A}bstract {M}eaning {R}epresentation in
  pre-trained language models.
\newblock In \emph{Proceedings of the 2nd Workshop on Deep Learning on Graphs
  for Natural Language Processing (DLG4NLP 2022)}.

\bibitem[{Cornia et~al.(2019)Cornia, Baraldi, and Cucchiara}]{cornia.etal.2019}
Marcella Cornia, Lorenzo Baraldi, and Rita Cucchiara. 2019.
\newblock Show, control, and tell: A framework for generating controllable and
  grounded captioning.
\newblock In \emph{Proceedings of the IEEE Conference on Computer Vision and
  Pattern Recognition}.

\bibitem[{Damodaran et~al.(2021)Damodaran, Chakravarthy, Kumar, Umapathy,
  Mitamura, Nakashima, Garcia, and Chu}]{damodaran2021understanding}
Vinay Damodaran, Sharanya Chakravarthy, Akshay Kumar, Anjana Umapathy, Teruko
  Mitamura, Yuta Nakashima, Noa Garcia, and Chenhui Chu. 2021.
\newblock Understanding the role of scene graphs in visual question answering.
\newblock \emph{arXiv preprint arXiv:2101.05479}.

\bibitem[{Denkowski and Lavie(2014)}]{denkowski2014meteor}
Michael Denkowski and Alon Lavie. 2014.
\newblock Meteor universal: Language specific translation evaluation for any
  target language.
\newblock In \emph{Workshop on Statistical Machine Translation}.

\bibitem[{Drozdov et~al.(2022)Drozdov, Zhou, Florian, McCallum, Naseem, Kim,
  and Astudillo}]{drozdov.etal.2022-amr}
Andrew Drozdov, Jiawei Zhou, Radu Florian, Andrew McCallum, Tahira Naseem, Yoon
  Kim, and Ramon~Fernandez Astudillo. 2022.
\newblock Inducing and using alignments for transition-based {AMR} parsing.
\newblock In \emph{North American Chapter of the Association for Computational
  Linguistics}.

\bibitem[{Fei-Fei et~al.(2007)Fei-Fei, Iyer, Koch, and Perona}]{fei2007we}
Li~Fei-Fei, Asha Iyer, Christof Koch, and Pietro Perona. 2007.
\newblock What do we perceive in a glance of a real-world scene?
\newblock \emph{Journal of Vision}, 7(1).

\bibitem[{Fellbaum(1998)}]{wordnet.1998}
Christiane Fellbaum, editor. 1998.
\newblock \emph{WordNet: An Electronic Lexical Database}.
\newblock Cambridge, MA: MIT Press.

\bibitem[{Gao et~al.(2018)Gao, Xiong, and Grauman}]{gao2018im2flow}
Ruohan Gao, Bo~Xiong, and Kristen Grauman. 2018.
\newblock Im2flow: Motion hallucination from static images for action
  recognition.
\newblock In \emph{Proceedings of the IEEE Conference on Computer Vision and
  Pattern Recognition}.

\bibitem[{He et~al.(2016)He, Zhang, Ren, and Sun}]{he.etal.2016-resnet}
Kaiming He, Xiangyu Zhang, Shaoqing Ren, and Jian Sun. 2016.
\newblock Deep residual learning for image recognition.
\newblock In \emph{Proceedings of the IEEE Conference on Computer Vision and
  Pattern Recognition}.

\bibitem[{Herath et~al.(2017)Herath, Harandi, and Porikli}]{herath2017going}
Samitha Herath, Mehrtash Harandi, and Fatih Porikli. 2017.
\newblock Going deeper into action recognition: A survey.
\newblock \emph{Image and vision computing}, 60.

\bibitem[{Hildebrandt et~al.(2020)Hildebrandt, Li, Koner, Tresp, and
  G{\"u}nnemann}]{hildebrandt2020scene}
Marcel Hildebrandt, Hang Li, Rajat Koner, Volker Tresp, and Stephan
  G{\"u}nnemann. 2020.
\newblock Scene graph reasoning for visual question answering.
\newblock \emph{arXiv preprint arXiv:2007.01072}.

\bibitem[{Hou et~al.(2020)Hou, Peng, Qiao, and Tao}]{hou2020visual}
Zhi Hou, Xiaojiang Peng, Yu~Qiao, and Dacheng Tao. 2020.
\newblock Visual compositional learning for human-object interaction detection.
\newblock In \emph{European Conference on Computer Vision}.

\bibitem[{Johnson et~al.(2015{\natexlab{a}})Johnson, Krishna, Stark, Li,
  Shamma, Bernstein, and Li}]{johnson.etal.2015}
J.~Johnson, R.~Krishna, M.~Stark, L.~J. Li, D.~A. Shamma, M.~S. Bernstein, and
  F.~F. Li. 2015{\natexlab{a}}.
\newblock Image retrieval using scene graphs.
\newblock In \emph{IEEE Conference on Computer Vision and Pattern Recognition}.

\bibitem[{Johnson et~al.(2015{\natexlab{b}})Johnson, Krishna, Stark, Li,
  Shamma, Bernstein, and Fei-Fei}]{johnson2015image}
Justin Johnson, Ranjay Krishna, Michael Stark, Li-Jia Li, David Shamma, Michael
  Bernstein, and Li~Fei-Fei. 2015{\natexlab{b}}.
\newblock Image retrieval using scene graphs.
\newblock In \emph{Proceedings of the IEEE conference on computer vision and
  pattern recognition}.

\bibitem[{{Karpathy} and {Fei-Fei}(2015)}]{Karpathy2015}
A.~{Karpathy} and L.~{Fei-Fei}. 2015.
\newblock Deep visual-semantic alignments for generating image descriptions.
\newblock In \emph{IEEE Conference on Computer Vision and Pattern Recognition}.

\bibitem[{Kong and Fu(2022)}]{kong2022human}
Yu~Kong and Yun Fu. 2022.
\newblock Human action recognition and prediction: A survey.
\newblock \emph{International Journal of Computer Vision}, 130(5).

\bibitem[{Krishna et~al.(2016)Krishna, Zhu, Groth, Johnson, Hata, Kravitz,
  Chen, Kalantidis, Li, Shamma, Bernstein, and Fei-Fei}]{krishnavisualgenome}
Ranjay Krishna, Yuke Zhu, Oliver Groth, Justin Johnson, Kenji Hata, Joshua
  Kravitz, Stephanie Chen, Yannis Kalantidis, Li-Jia Li, David~A Shamma,
  Michael Bernstein, and Li~Fei-Fei. 2016.
\newblock Visual genome: Connecting language and vision using crowdsourced
  dense image annotations.

\bibitem[{Lee et~al.(2018)Lee, Chen, Hua, Hu, and He}]{lee2018stacked}
Kuang-Huei Lee, Xi~Chen, Gang Hua, Houdong Hu, and Xiaodong He. 2018.
\newblock Stacked cross attention for image-text matching.
\newblock In \emph{European Conference on Computer Vision}.

\bibitem[{Lewis et~al.(2020)Lewis, Liu, Goyal, Ghazvininejad, Mohamed, Levy,
  Stoyanov, and Zettlemoyer}]{lewis.etal.2020-bart}
M.~Lewis, Y.~Liu, N.~Goyal, M.~Ghazvininejad, A.~Mohamed, O.~Levy, V.~Stoyanov,
  and L.~Zettlemoyer. 2020.
\newblock {BART}: Denoising sequence-to-sequence pre-training for natural
  language generation, translation, and comprehension.
\newblock In \emph{Association for Computational Linguistics}.

\bibitem[{Li and Jiang(2019)}]{li2019know}
Xiangyang Li and Shuqiang Jiang. 2019.
\newblock Know more say less: Image captioning based on scene graphs.
\newblock \emph{IEEE Transactions on Multimedia}, 21(8):2117--2130.

\bibitem[{Li et~al.(2018)Li, Fang, Yang, Wang, Lu, and Yang}]{li2018flow}
Yijun Li, Chen Fang, Jimei Yang, Zhaowen Wang, Xin Lu, and Ming-Hsuan Yang.
  2018.
\newblock Flow-grounded spatial-temporal video prediction from still images.
\newblock In \emph{Proceedings of the European Conference on Computer Vision}.

\bibitem[{Liao et~al.(2018)Liao, Lebanoff, and Liu}]{liao.etal.2018}
Kexin Liao, Logan Lebanoff, and Fei Liu. 2018.
\newblock {A}bstract {M}eaning {R}epresentation for multi-document
  summarization.
\newblock In \emph{the 27th International Conference on Computational
  Linguistics}.

\bibitem[{Lin et~al.(2014)Lin, Maire, Belongie, Hays, Perona, Ramanan,
  Doll{\'a}r, and Zitnick}]{Lin2014}
Tsung-Yi Lin, Michael Maire, Serge Belongie, James Hays, Pietro Perona, Deva
  Ramanan, Piotr Doll{\'a}r, and C.~Lawrence Zitnick. 2014.
\newblock {Microsoft COCO}: Common objects in context.
\newblock In \emph{European Conference on Computer Vision}.

\bibitem[{Liu et~al.(2015)Liu, Flanigan, Thomson, Sadeh, and
  Smith}]{liu.etal.2015}
Fei Liu, Jeffrey Flanigan, Sam Thomson, Norman Sadeh, and Noah~A. Smith. 2015.
\newblock Toward abstractive summarization using semantic representations.
\newblock In \emph{North American Chapter of the Association for Computational
  Linguistics}.

\bibitem[{Liu et~al.(2020{\natexlab{a}})Liu, Ouyang, Wang, Fieguth, Chen, Liu,
  and Pietik{\"a}inen}]{liu2020deep}
Li~Liu, Wanli Ouyang, Xiaogang Wang, Paul Fieguth, Jie Chen, Xinwang Liu, and
  Matti Pietik{\"a}inen. 2020{\natexlab{a}}.
\newblock Deep learning for generic object detection: A survey.
\newblock \emph{International Journal of Computer Vision}, 128(2).

\bibitem[{Liu et~al.(2020{\natexlab{b}})Liu, Jiang, He, Chen, Liu, Gao, and
  Han}]{liu.etal.2020-radam}
Liyuan Liu, Haoming Jiang, Pengcheng He, Weizhu Chen, Xiaodong Liu, Jianfeng
  Gao, and Jiawei Han. 2020{\natexlab{b}}.
\newblock On the variance of the adaptive learning rate and beyond.
\newblock In \emph{International Conference on Learning Representations}.

\bibitem[{Liu et~al.(2016)Liu, Anguelov, Erhan, Szegedy, Reed, Fu, and
  Berg}]{liu2016ssd}
Wei Liu, Dragomir Anguelov, Dumitru Erhan, Christian Szegedy, Scott Reed,
  Cheng-Yang Fu, and Alexander~C Berg. 2016.
\newblock Ssd: Single shot multibox detector.
\newblock In \emph{European Conference on Computer Vision}.

\bibitem[{Lu et~al.(2016)Lu, Krishna, Bernstein, and Fei-Fei}]{lu2016visual}
Cewu Lu, Ranjay Krishna, Michael Bernstein, and Li~Fei-Fei. 2016.
\newblock Visual relationship detection with language priors.
\newblock In \emph{European Conference on Computer Vision}.

\bibitem[{Markman(1990)}]{markman.1990}
Ellen~M. Markman. 1990.
\newblock Constraints children place on word meanings.
\newblock \emph{Cognitive Science}, 14.

\bibitem[{M{\`a}rquez et~al.(2008)M{\`a}rquez, Carreras, Litkowski, and
  Stevenson}]{marquez-etal-2008-special}
Llu{\'\i}s M{\`a}rquez, Xavier Carreras, Kenneth~C. Litkowski, and Suzanne
  Stevenson. 2008.
\newblock Special issue introduction: Semantic role labeling: An introduction
  to the special issue.
\newblock \emph{Computational Linguistics}, 34(2).

\bibitem[{Naseem et~al.(2021)Naseem, Blodgett, Kumaravel, O'Gorman, Lee,
  Flanigan, Astudillo, Florian, Roukos, and Schneider}]{naseem.etal.2021}
Tahira Naseem, Austin Blodgett, Sadhana Kumaravel, Tim O'Gorman, Young-Suk Lee,
  Jeffrey Flanigan, Ram{\'{o}}n~Fernandez Astudillo, Radu Florian, Salim
  Roukos, and Nathan Schneider. 2021.
\newblock {DocAMR}: Multi-sentence {AMR} representation and evaluation.
\newblock In \emph{arXiv}.

\bibitem[{Papineni et~al.(2002)Papineni, Roukos, Ward, and
  Zhu}]{papineni2002bleu}
Kishore Papineni, Salim Roukos, Todd Ward, and Wei-Jing Zhu. 2002.
\newblock {B}leu: a method for automatic evaluation of machine translation.
\newblock In \emph{40th Annual Meeting of the Association for Computational
  Linguistics}.

\bibitem[{Ren et~al.(2015)Ren, He, Girshick, and Sun}]{ren2015faster}
Shaoqing Ren, Kaiming He, Ross Girshick, and Jian Sun. 2015.
\newblock Faster {R-CNN}: Towards real-time object detection with region
  proposal networks.
\newblock \emph{Advances in Neural Information Processing Systems}, 28.

\bibitem[{Ruppenhofer et~al.(2016)Ruppenhofer, Michael~Ellsworth, Johnson,
  Baker, and Scheffczyk}]{framenet}
Josef Ruppenhofer, Miriam R. L~Petruck Michael~Ellsworth, Christopher~R.
  Johnson, Collin~F. Baker, and Jan Scheffczyk. 2016.
\newblock \emph{FrameNet II: Extended Theory and Practice}.

\bibitem[{Schroeder and Tripathi(2020)}]{schroeder2020structured}
Brigit Schroeder and Subarna Tripathi. 2020.
\newblock Structured query-based image retrieval using scene graphs.
\newblock In \emph{Proceedings of the IEEE/CVF Conference on Computer Vision
  and Pattern Recognition Workshops}, pages 178--179.

\bibitem[{Schuster et~al.(2015)Schuster, Krishna, Chang, Fei-Fei, and
  Manning}]{schuster2015generating}
Sebastian Schuster, Ranjay Krishna, Angel Chang, Li~Fei-Fei, and Christopher~D
  Manning. 2015.
\newblock Generating semantically precise scene graphs from textual
  descriptions for improved image retrieval.
\newblock In \emph{Proceedings of the fourth workshop on vision and language},
  pages 70--80.

\bibitem[{Song and Gildea(2019)}]{song.gildea.2019-sembleu}
Linfeng Song and Daniel Gildea. 2019.
\newblock {S}em{B}leu: A robust metric for {AMR} parsing evaluation.
\newblock In \emph{57th Annual Meeting of the Association for Computational
  Linguistics}.

\bibitem[{Vedantam et~al.(2015)Vedantam, Zitnick, and
  Parikh}]{vedantam2015cider}
Ramakrishna Vedantam, C.~Lawrence Zitnick, and Devi Parikh. 2015.
\newblock Cider: Consensus-based image description evaluation.
\newblock In \emph{IEEE Conference on Computer Vision and Pattern Recognition}.

\bibitem[{Wang et~al.(2020)Wang, Wang, Yao, Shan, and Chen}]{wang2020cross}
Sijin Wang, Ruiping Wang, Ziwei Yao, Shiguang Shan, and Xilin Chen. 2020.
\newblock Cross-modal scene graph matching for relationship-aware image-text
  retrieval.
\newblock In \emph{Proceedings of the IEEE/CVF winter conference on
  applications of computer vision}, pages 1508--1517.

\bibitem[{Xia et~al.(2021)Xia, Li, Wang, and Zhang}]{xia.etal.2021-stackedamr}
Qingrong Xia, Zhenghua Li, Rui Wang, and Min Zhang. 2021.
\newblock Stacked {AMR} parsing with silver data.
\newblock In \emph{Findings of the Association for Computational Linguistics:
  EMNLP}.

\bibitem[{Yang et~al.(2019)Yang, Tang, Zhang, and Cai}]{yang2019auto}
Xu~Yang, Kaihua Tang, Hanwang Zhang, and Jianfei Cai. 2019.
\newblock Auto-encoding scene graphs for image captioning.
\newblock In \emph{Proceedings of the IEEE/CVF Conference on Computer Vision
  and Pattern Recognition}, pages 10685--10694.

\bibitem[{Yatskar et~al.(2016)Yatskar, Zettlemoyer, and
  Farhadi}]{yatskar.etal.2016}
Mark Yatskar, Luke Zettlemoyer, and Ali Farhadi. 2016.
\newblock Situation recognition: Visual semantic role labeling for image
  understanding.
\newblock In \emph{IEEE Conference on Computer Vision and Pattern Recognition}.

\bibitem[{Zellers et~al.(2018)Zellers, Yatskar, Thomson, and
  Choi}]{zellers.etal.2018}
Rowan Zellers, Mark Yatskar, Sam Thomson, and Yejin Choi. 2018.
\newblock Neural motifs: Scene graph parsing with global context.
\newblock In \emph{IEEE Conference on Computer Vision and Pattern Recognition}.

\bibitem[{Zhang et~al.(2019{\natexlab{a}})Zhang, Chao, and
  Xuan}]{zhang2019empirical}
Cheng Zhang, Wei-Lun Chao, and Dong Xuan. 2019{\natexlab{a}}.
\newblock An empirical study on leveraging scene graphs for visual question
  answering.
\newblock \emph{arXiv preprint arXiv:1907.12133}.

\bibitem[{Zhang et~al.(2022)Zhang, Song, Li, Zhou, and
  Song}]{zhang.etal.2022-survey}
Hanqing Zhang, Haolin Song, Shaoyu Li, Ming Zhou, and Dawei Song. 2022.
\newblock A survey of controllable text generation using transformer-based
  pre-trained language models.
\newblock \emph{arXiv: \url{https://arxiv.org/abs/2201.05337}}.

\bibitem[{Zhang et~al.(2017)Zhang, Kyaw, Chang, and Chua}]{zhang2017visual}
Hanwang Zhang, Zawlin Kyaw, Shih-Fu Chang, and Tat-Seng Chua. 2017.
\newblock Visual translation embedding network for visual relation detection.
\newblock In \emph{Proceedings of the IEEE Conference on Computer Vision and
  Pattern Recognition}.

\bibitem[{Zhang et~al.(2019{\natexlab{b}})Zhang, Ma, Duh, and
  Van~Durme}]{zhang.etal.2019-amr}
Sheng Zhang, Xutai Ma, Kevin Duh, and Benjamin Van~Durme. 2019{\natexlab{b}}.
\newblock {AMR} parsing as sequence-to-graph transduction.
\newblock In \emph{Proceedings of the 57th Annual Meeting of the Association
  for Computational Linguistics}.

\bibitem[{Zhong et~al.(2020)Zhong, Wang, Chen, Yu, and
  Li}]{zhong2020comprehensive}
Yiwu Zhong, Liwei Wang, Jianshu Chen, Dong Yu, and Yin Li. 2020.
\newblock Comprehensive image captioning via scene graph decomposition.
\newblock In \emph{European Conference on Computer Vision}, pages 211--229.
  Springer.

\bibitem[{Zhu et~al.(2022)Zhu, Zhang, Jiang, Dang, Hou, Shen, Feng, Zhao, Miao,
  Shah et~al.}]{zhu.etal.2022.survey}
Guangming Zhu, Liang Zhang, Youliang Jiang, Yixuan Dang, Haoran Hou, Peiyi
  Shen, Mingtao Feng, Xia Zhao, Qiguang Miao, Syed Afaq~Ali Shah, et~al. 2022.
\newblock Scene graph generation: A comprehensive survey.
\newblock \emph{arXiv preprint arXiv:2201.00443}.

\bibitem[{Zou et~al.(2021)Zou, Wang, Hu, Liu, Wu, Zhao, Li, Zhang, Zhang, Wei
  et~al.}]{zou2021end}
Cheng Zou, Bohan Wang, Yue Hu, Junqi Liu, Qian Wu, Yu~Zhao, Boxun Li, Chenguang
  Zhang, Chi Zhang, Yichen Wei, et~al. 2021.
\newblock End-to-end human object interaction detection with hoi transformer.
\newblock In \emph{Proceedings of the IEEE/CVF conference on computer vision
  and pattern recognition}.

\end{thebibliography}
\bibliographystyle{acl_natbib}

\appendix
\clearpage

\section{AMR vs. SG: Entity and Relation Categorization Details}
\label{sec:appendix}

% \usepackage[normalem]{ulem}

%\begin{table*}[t]
%\centering
%\begin{tabular}{@{}lc|ccc@{}}
%\cmidrule{1-5}

The analysis provided in Section~\ref{s:analysis} requires us to annotate the entities and relations of a sample of AMRs and SGs into a pre-defined set of categories.
We first select all images that appear in both MSCOCO~\cite{Lin2014} and Visual Genome, so we have access to ground-truth scene graphs, as well as captions from which we can generate AMR graphs for the same set of images. We use a single AMR per image, generated from the longest caption, but include all SGs associated with an image in our analysis.
For each SG and AMR graph, we consider the entities and relations corresponding to ${\sim}900$ most frequent types (around $1.3$M entity and $1$M relation instances for SGs; and around $130$K entity and $150$K relation instances for AMRs). 
% \todo{Kalliopi: add number of instances}
We annotate these into a pre-defined set of entity and relation categories, including those defined by~\cite{zellers.etal.2018} plus a few we add to cover new AMR relations. Table~\ref{tab:analysis} provides a breakdown of the categories,
as well as examples of word types we considered to belong to each category. The table also provides the total number of word types per category and percentages of instances across all types for each category.

Next, we describe our annotation process. SG nodes (entities) come with their most common WordNet sense annotations, which we use to identify their categories. For SG relations, we manually annotate their categories.
To annotate AMR entities and relations, we follow a similar procedure, by automatically finding the most common WordNet sense for non-predicate AMR nodes (assuming most of these will be entities) and correcting them if needed. For example, the automatically-identified most common sense of {\it mouse} is the Animal sense, whereas in our captions, almost all instances of the word point to the computer mouse (Artifact).
For any remaining concepts, including predicate nodes (e.g., {\it eat}, {\it stand}) and entities for which a category cannot be assigned automatically, we manually identify their categories.

%%%%%%
\section{Distribution of AMR Node Types}\label{a:amr_node_dist}

\begin{figure}[h!]
    \centering
    \includegraphics[width=\linewidth]{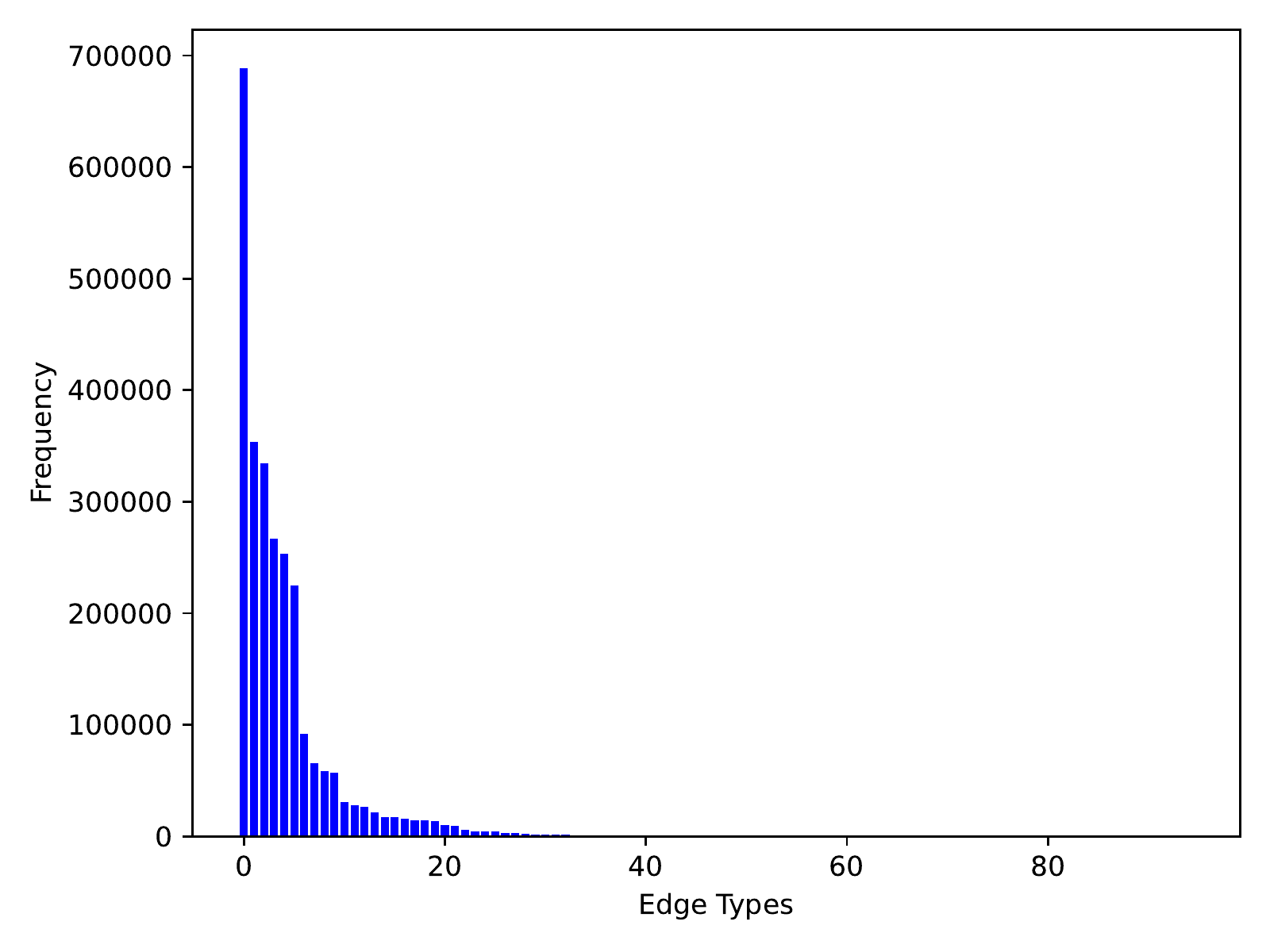}
    \caption{Frequency of the $90$ AMR role/edge types prior to the refinement process, which exhibits the characteristics of a long-tail distribution.}
    \label{fig:edge_type_stats}
\end{figure}

Fig.~\ref{fig:edge_type_stats} shows the distribution of the $90$ AMR role/edge types in our training data. As we can see, keeping the top-$20$ types is justified given the skewed distribution of the types. Future work will need to examine the nature of the less frequent relations, and the implications of removing them from AMR graphs.

%%%%%%
\section{Meta-AMR Construction Example}\label{a:meta_amr_example}

\begin{figure*}[b]
    \centering
    \includegraphics[width=0.9\linewidth]{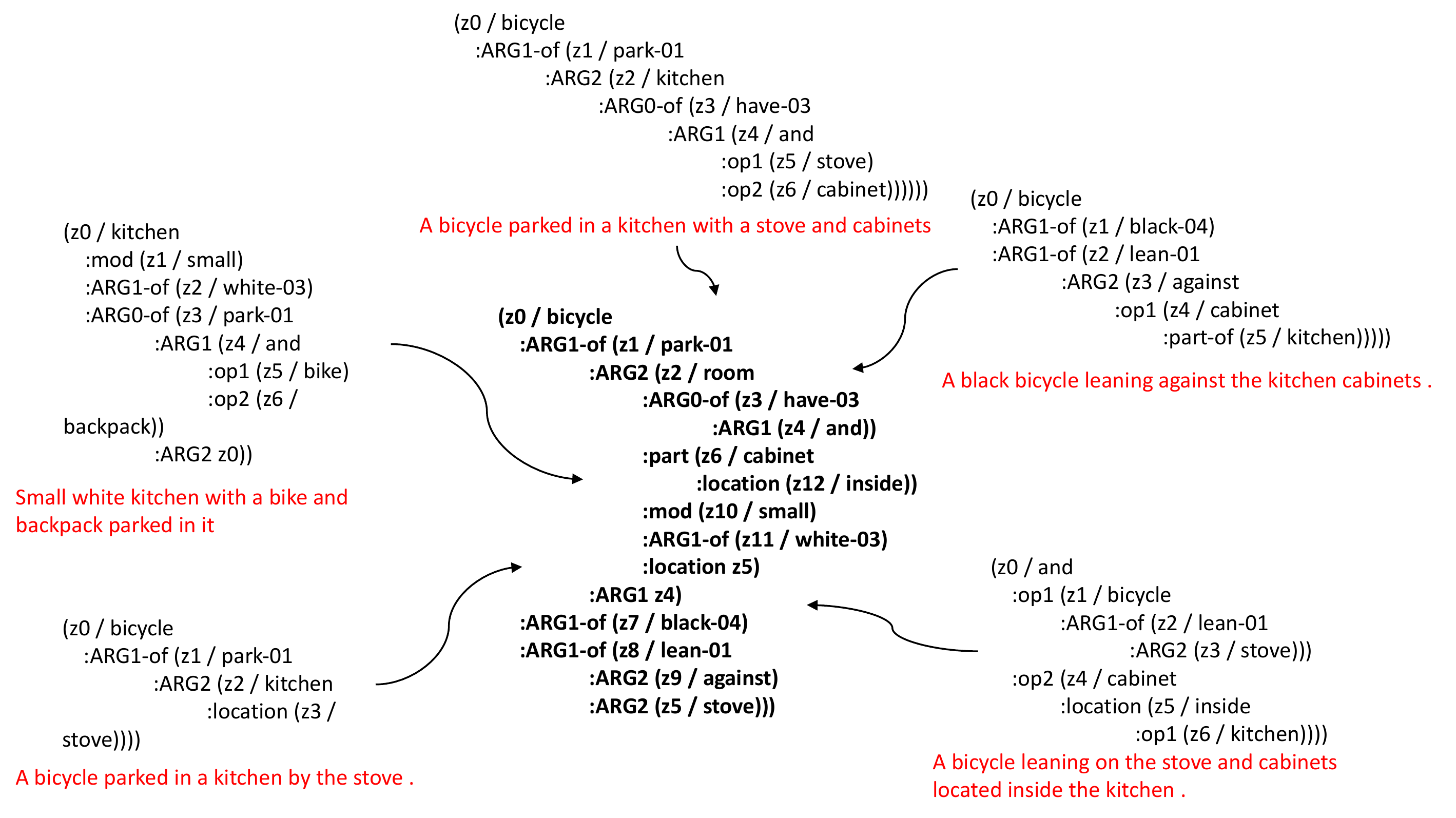}
    \caption{An example of five caption AMRs and their corresponding meta-AMR. Captions are marked as red.}
    \label{fig:metaAMR-example}
\end{figure*}

Fig.~\ref{fig:metaAMR-example} shows an example of how a meta-AMR is constructed from five caption-level AMRs. The corresponding captions are provided in red, and the AMR graphs are given in PENMAN notation.

%%%%%%
\section{Ablations}\label{ss:ablations}

\paragraph{Effect of input on node prediction performance.}
Table~\ref{res:node-abl} presents performance of meta-AMR node prediction (first stage of \textsc{Img2Amr}$_\mathrm{2stage}$) with different input combinations, in terms of Precision and Recall (when predicted and ground-truth nodes are taken as sets), and \textsc{Bleu}-1 (when the order of nodes in the final linearized AMR is taken into consideration).
These results suggest that an overall best performance is achieved by using all input features, namely regions, tags and global image feature. %Our node prediction stage thus always uses this combination of input.

\begin{table}[h!]
    \centering\small
    \begin{tabular}{@{}ccc|ccc@{}}
    \toprule
    \{$\bf{r}$\} & \{$\bf{q}$\}  & \{$\bf{g}$\} & {\small Recall} & {\small Precision}     & {\small \textsc{Bleu}-1}      \\ \midrule
    \checkmark & -  & -  & 34.5          & 47.1          & 33.1         \\
    -  & \checkmark & -  & 30.4          & 42.8          & 29.7          \\
    -  & -  & \checkmark & 30.6          & 39.9          & 29.1         \\
    \checkmark & \checkmark & -  & {\ul 35.8}    & \textbf{49.0} & {\ul 34.3}  \\
    \checkmark & -  & \checkmark & 35.1          & 47.5          & 33.9    \\
    -  & \checkmark & \checkmark & 32.9          & 46.5          & 32.1       \\
    \checkmark & \checkmark & \checkmark & \textbf{36.7} & {\ul 48.4}    & \textbf{35.6}    \\ \bottomrule
    \end{tabular}
    \caption{\textsc{val} performance of meta-AMR node prediction (first stage of \textsc{Img2Amr}$_\mathrm{2stage}$) with different input combinations.}
    \label{res:node-abl}
\end{table}

\paragraph{Effect of input on parsing performance.}
We train our \textsc{Img2Amr} models with different inputs to the encoders, and evaluate on \textsc{val} set. Specifically, the input to the model may contain the global image feature $\bf{g}$, region embeddings $\bf{r}$, tag embeddings $\bf{q}$ (for the first encoder), and node embeddings $\bf{n}$ (for the second encoder of \textsc{Img2Amr}$_\mathrm{2stage}$).
%\hl{VP: are you using the same notation as that in Fig.1?  It looks different to me.}
Table~\ref{res:metaAMR-abl} reports the \textsc{val} results of our two models (trained and tested with meta-AMRs) with different input combinations (region embeddings, tag embeddings, global image features) for the $\mathrm{direct}$ model, and (node embeddings, global image features, region embeddings) for the second encoder of the $\mathrm{2stage}$ model.
For \textsc{Img2Amr}$_\mathrm{2stage}$, we fix the input of the first encoder to the best combination according to Table~\ref{res:node-abl} above, and ablate over the input of the second encoder. Both models are trained and tested with meta-AMRs. As we can see, richer input generally results in better performance. We can also see a big drop in the performance of \textsc{Img2Amr}$_\mathrm{direct}$ when only region features are used as input, suggesting that tags can help associate mappings between regions and AMR concepts. 
%\hl{More details in last row because img2amrdirect with \{r\} has substantially lower performance, I believe it refers to the comparison of the 2nd + 3rd row of Img2amrdirect vs all cases of img2amr2stage.}

\begin{table}[h!]
\centering\small
\begin{tabular}{@{}c|ccc@{}}
\toprule
  {\small Model Input}  & {\small \textsc{Smatch}}  & {\small \textsc{SemBleu}-1} & {\small \textsc{SemBleu}-2}    \\ \midrule 
%\multirow{3}{*}{\rotatebox[origin=c]{90}{\STAB{$\mathrm{direct}$}}}
\multicolumn{4}{c}{\textsc{Img2Amr}$_\mathrm{direct}$} \\ \cmidrule{1-4}
 \{$\bf{r}$\}                  & 30.3             &      18.6        & 5.4            \\
 \{$\bf{r}, \bf{q}$\}          & \textbf{39.1}  & 32.9          & 16.2          \\
 \{$\bf{r}, \bf{q}, \bf{g}$\}  & 39.0 & \textbf{33.7} & \textbf{16.4}          \\ \cmidrule{1-4} \morecmidrules \cmidrule{1-4} 
%\multirow{3}{*}{\rotatebox[origin=c]{90}{\STAB{$\mathrm{2stage}$}}} 
\multicolumn{4}{c}{\textsc{Img2Amr}$_\mathrm{2stage}$} \\ \cmidrule{1-4}
 \{$\bf{n}$\}                 & 39.3          & 31.3          & 16.1          \\
 \{$\bf{n}, \bf{g}$\}           & 39.6    & 31.9          &  16.3    \\
 \{$\bf{n}, \bf{g}, \bf{r}$\} & \textbf{40.4} & \textbf{32.6}    & \textbf{16.9} \\ \cmidrule{1-4} 
\end{tabular}
\caption{Ablation over model inputs on \textsc{val}, for both \textsc{Img2Amr} models. For \textsc{Img2Amr}$_\mathrm{2stage}$ we use all features \{$\bf{r}, \bf{q}, \bf{g}$\} as the 1st encoder input.
}
\label{res:metaAMR-abl}
\end{table}

\clearpage

\begin{table*}[t]
\centering\small
\begin{tabular}{@{}ll|cccc@{}}
\cmidrule{1-6}
&& \multicolumn{2}{c}{\#Types} & \multicolumn{2}{c}{\%Tokens}              \\ \cmidrule{3-6}
Cateogry                 & Example Types per Category & AMR  & SG  & AMR                 & SG                   \\ \cmidrule{1-6}
\multicolumn{6}{c}{\textsc{Entities}}   \\ \cmidrule{1-6}
Artifact             & clock, umbrella, bottle  & 128          & 128          & 22.7 & 24.4 \\
Part                 & eyes, finger, wing       & 21           & 44           & 3.1   & 13.1 \\
Location             & beach, mountain, kitchen & 86           & 52           & 20.7 & 11.2 \\
Person               & man, women, speaker      & 30           & 19           & 17.9 & 11\\
Flora/Nature         & ocean, tree, flower      & 20           & 34           & 6.1   & 10.2 \\
Clothing             & dress, scarf, suit       & 11           & 31           & 1.1   & 7.7  \\ 
Food                 & orange, donut, bread     & 52           & 23           & 8  & 2.8   \\
Animal               & horse, bird, cat         & 16           & 20           & 6.4   & 4.7   \\
Vehicle              & car, motorcycle, bicycle & 18           & 17           & 6.1   & 4.5   \\
Furniture            & table, chair, couch      & 9            & 10           & 4.0   & 2.9   \\
Structure            & window, tower, circle    & 13           & 18           & 2.1   & 5.4   \\
Building             & brick, house, cement     & 6            & 6            & 1.8   & 2.1   \\
\cmidrule{1-6}
\multicolumn{6}{c}{\textsc{Relations}}   \\\cmidrule{1-6}
Geometric            & down, edge, between      & 48           & 122          & 12.4 & 56.6 \\
Possessive           & have, wear, contain      & 5            & 42           & 5.9   & 30.6 \\
Semantic             & attempt, carry, eat      & 183          & 275          & 38.3 & 11.6 \\
Attribute Color      & color, white, blue       & 13           & 8            & 5.6   & 0.1   \\
Attribute            & young, small, colorful   & 82           & -            & 12.8 & -                    \\
AMR specific         & and, or, date-entity     & 8            & -            & 11.1 & -                    \\
Quantifier           & more, both, few          & 31           & 1            & 9.3  & 0.1    \\
Event                & soccer, party, festival  & 14           & -            & 3.4   & -                    \\
Misc                 & they, something, you     & 6            & 13           & 1.1   & 1.0 \\\cmidrule{1-6}
\end{tabular}
    \caption{The list of AMR and SG entity and relation categories, as well as examples of word types, number of types, and percentage of tokens per category.}
    \label{tab:analysis}
\end{table*}

\clearpage
\onecolumn
\section{Generated AMRs for the Qualitative Samples}\label{ss:qualitative_amrs}
\begin{figure*}[!hb]
\centering
\captionsetup{justification=centering}
    \subfigure[A couple of giraffe standing next to each other in a field near rocks walking in grass in a grassy area.]{%
      \includegraphics[clip, width=0.35\textwidth, height=0.25\textwidth]{figures/qualitative/2.jpeg}
    \hspace{0.03\textwidth}
      \includegraphics[clip, width=0.6\textwidth, height=0.25\textwidth]{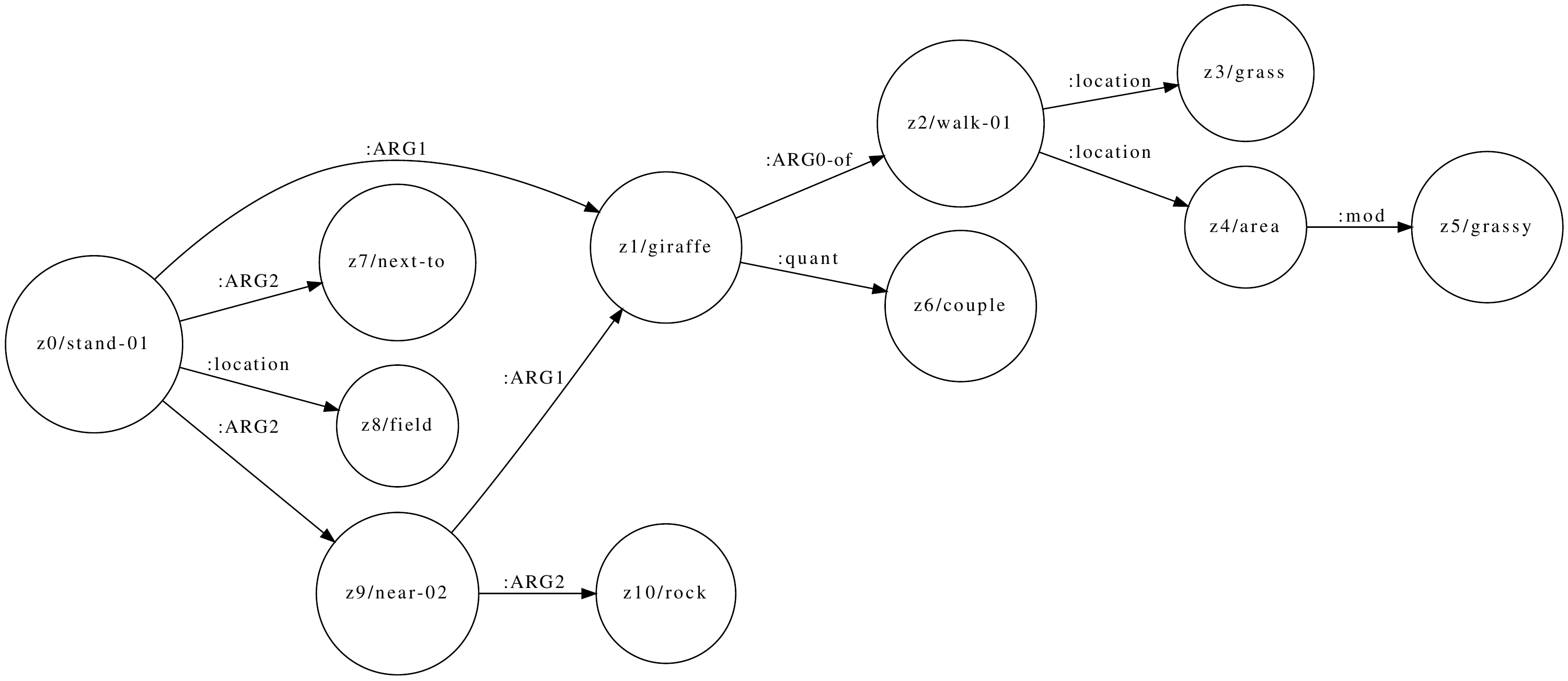}
    }
    \subfigure[A yellow and blue fire hydrant on a city street in front at an intersection sitting on the side of the road near a traffic position.]{%
      \includegraphics[clip, width=0.35\textwidth, height=0.45\textwidth]{figures/qualitative/3.jpeg}
      \hspace{0.03\textwidth}
      \includegraphics[clip, width=0.6\textwidth, height=0.45\textwidth]{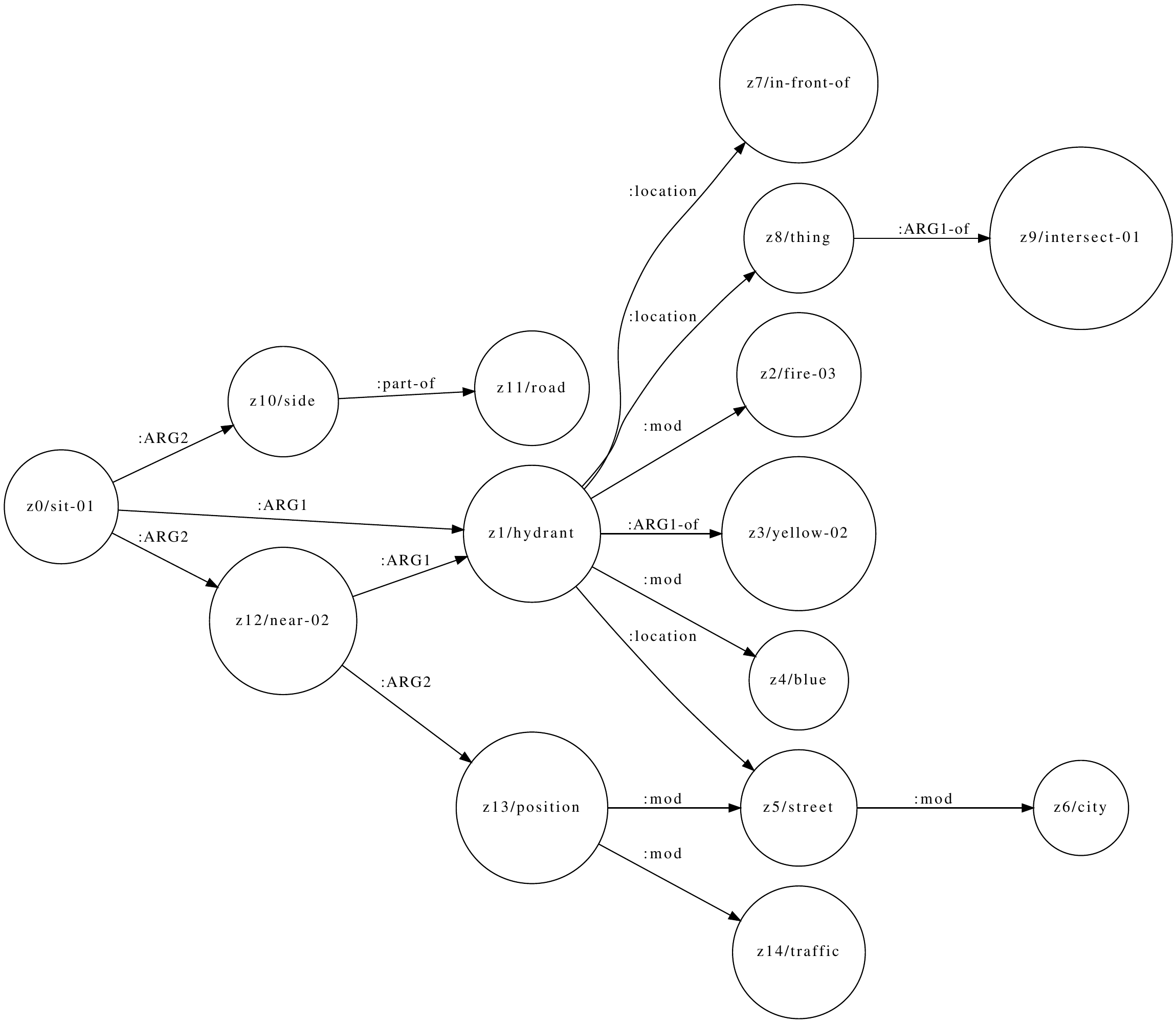}
    }
    \subfigure[A large long passenger train going across a wooden beach plate, traveling and passing by water.]{%
      \includegraphics[clip, width=0.35\textwidth, height=0.45\textwidth]{figures/qualitative/4.jpeg}
      \hspace{0.03\textwidth}
      \includegraphics[clip, width=0.6\textwidth, height=0.45\textwidth]{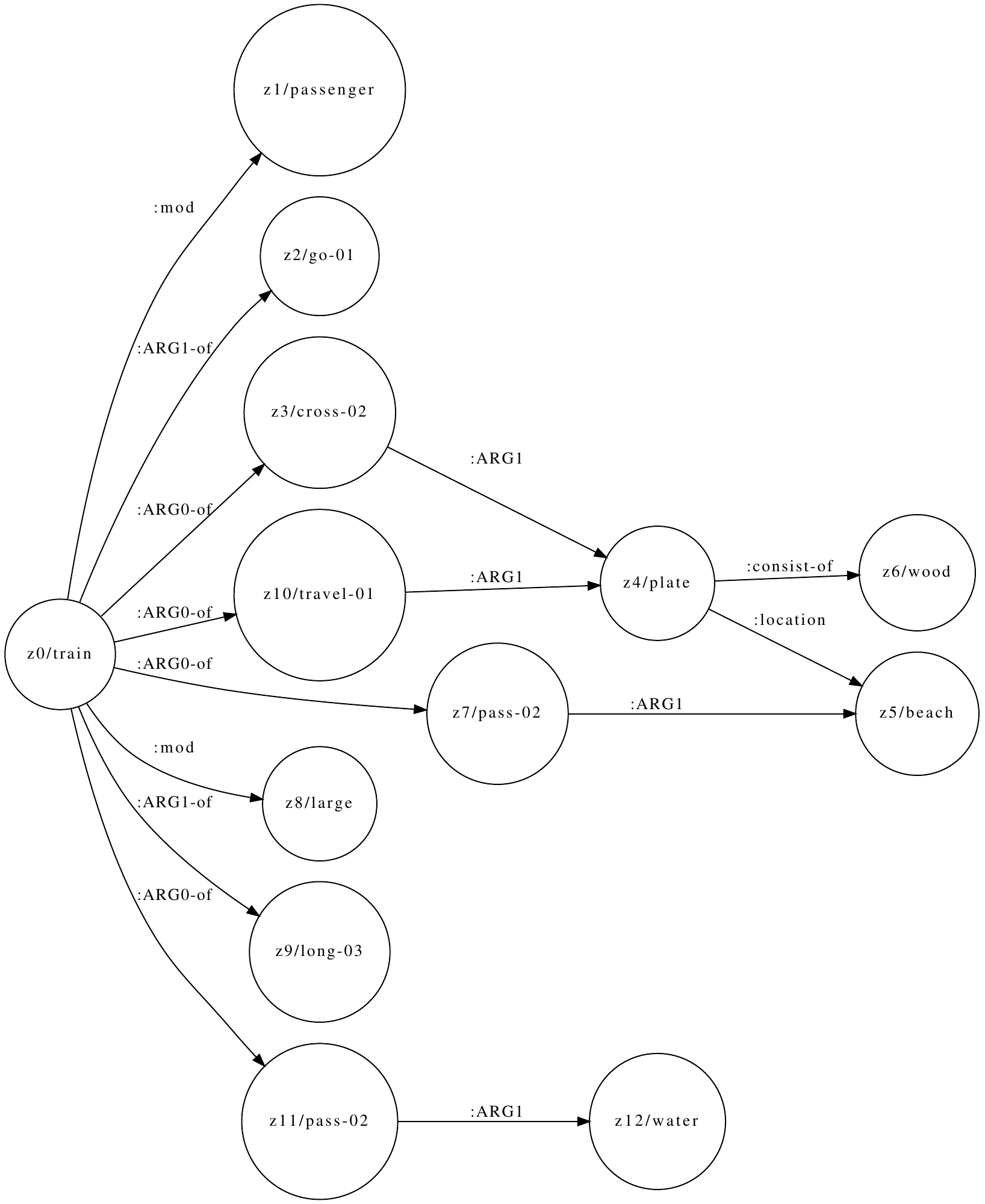}
    }
    \caption{Images used in Section~\ref{ss:qualitative}, along with their predicted AMRs and generated captions. Refer to Section~\ref{ss:qualitative} for more details.}
    \label{fig:qualitative_amrs_1}
\end{figure*}

\begin{figure*}[!hb]
\centering
\captionsetup{justification=centering}
    \subfigure[A woman sitting at a table eating a sandwich and holding a hot dog in a building smiling while eating.]{%
      \includegraphics[clip, width=0.35\textwidth, height=0.35\textwidth]{figures/qualitative/6.jpeg}
      \hspace{0.03\textwidth}
      \includegraphics[clip, width=0.6\textwidth, height=0.35\textwidth]{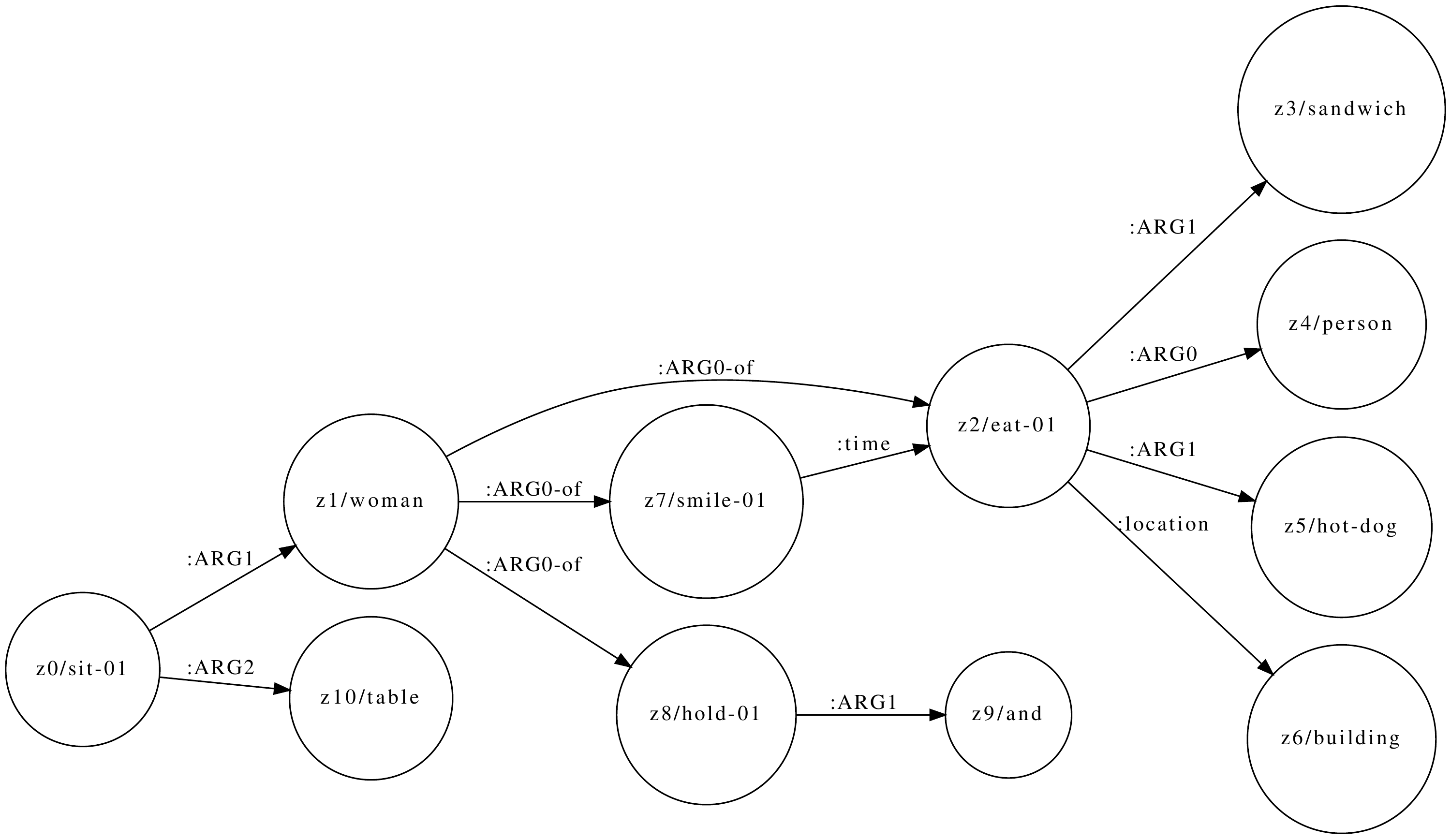}
    }
    \subfigure[A white area filled with lots of different kinds of donuts with various toppings sitting on them.]{%
      \includegraphics[clip, width=0.35\textwidth, height=0.35\textwidth]{figures/qualitative/7.jpeg}
      \hspace{0.03\textwidth}
      \includegraphics[clip, width=0.6\textwidth, height=0.35\textwidth]{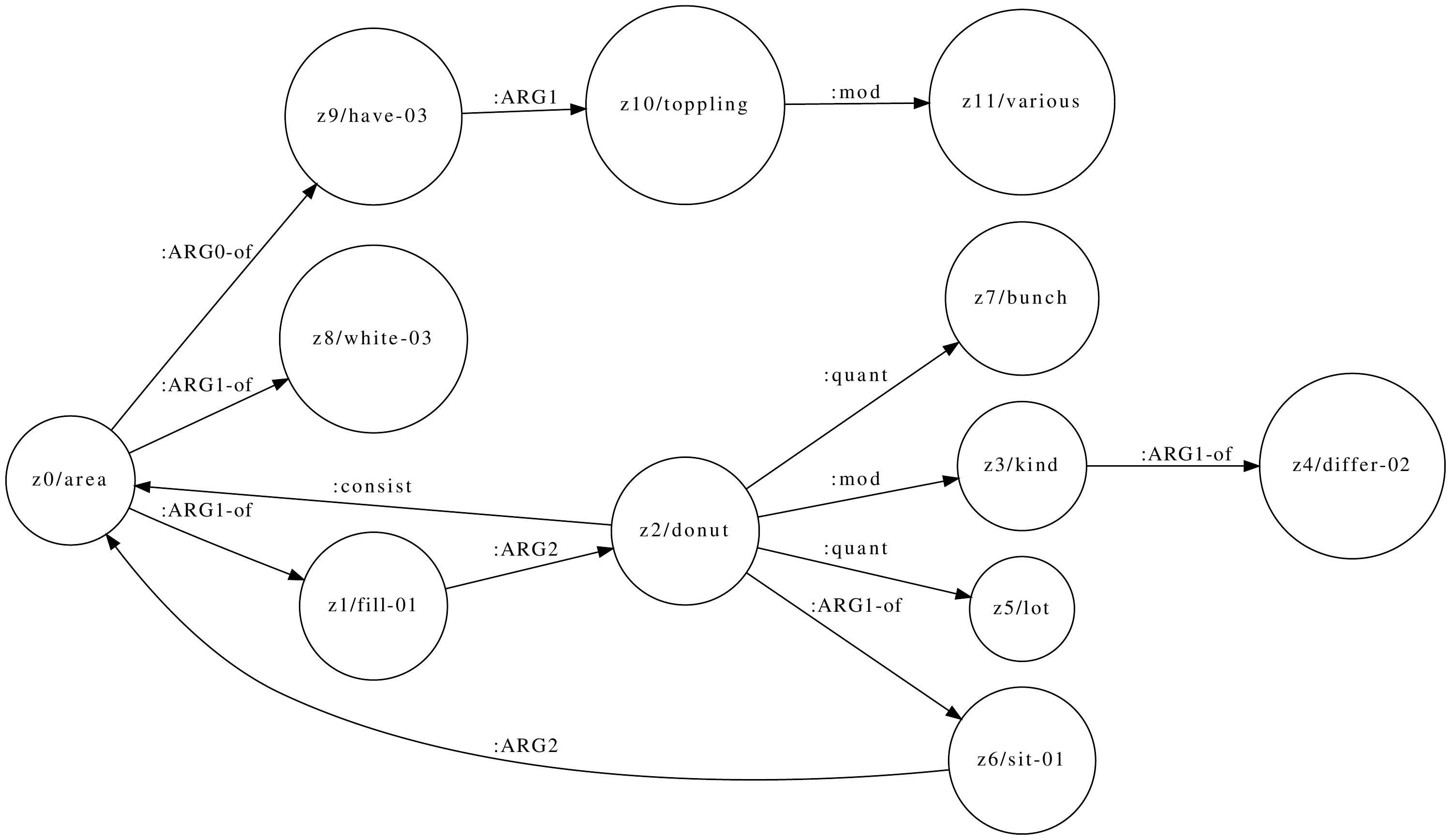}
    }
    \subfigure[A group of people sitting around at a dining table with water posing for a picture.]{%
      \includegraphics[clip, width=0.35\textwidth, height=0.35\textwidth]{figures/qualitative/8.jpeg}
      \hspace{0.03\textwidth}
      \includegraphics[clip, width=0.6\textwidth, height=0.35\textwidth]{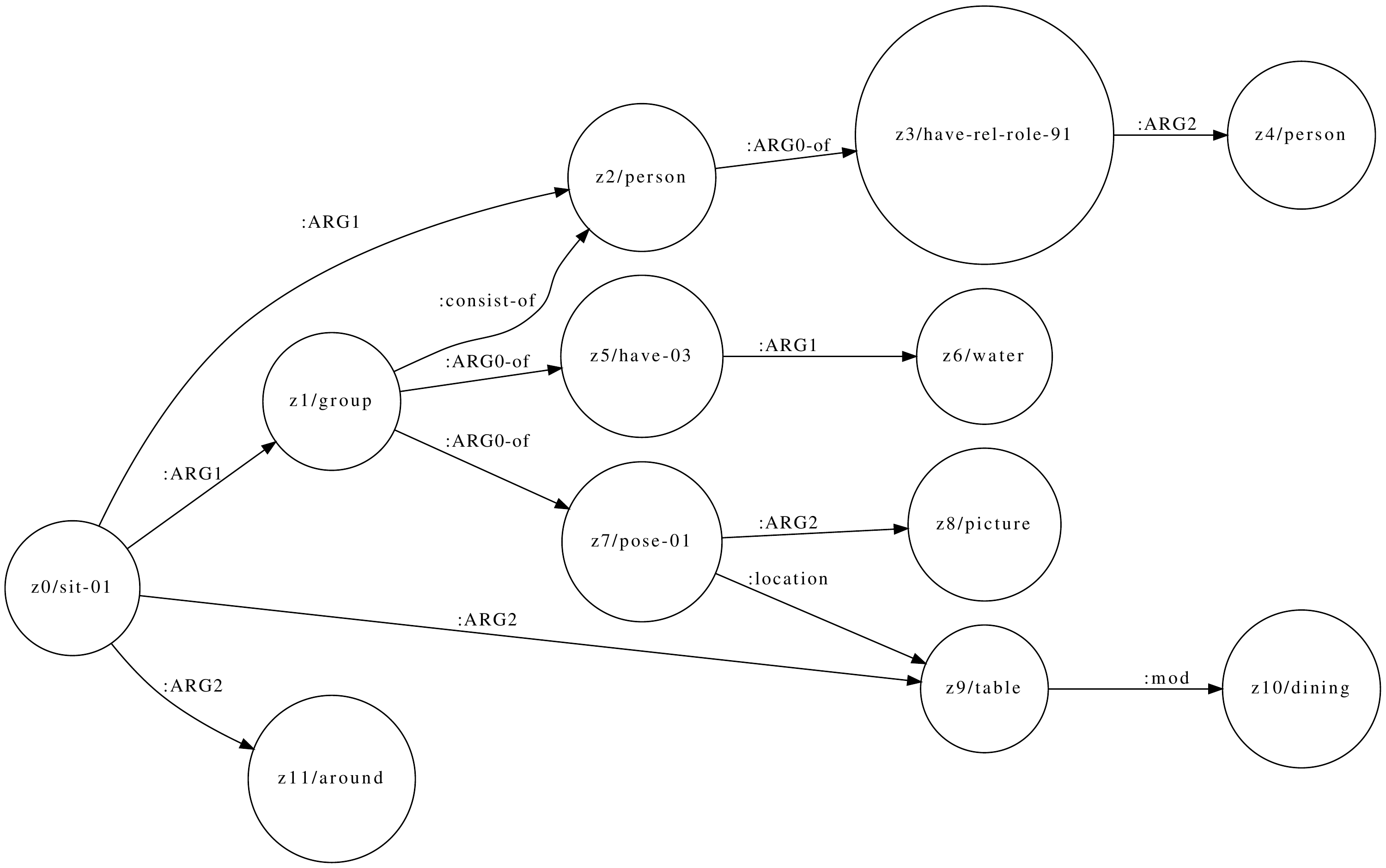}
    }
    \caption{(cont) Images used in Section~\ref{ss:qualitative}, along with their predicted AMRs and generated captions. Refer to Section~\ref{ss:qualitative} for more details.}
    \label{fig:qualitative_amrs_2}
\end{figure*}

\begin{figure*}[!hb]
\centering
\captionsetup{justification=centering}
    \subfigure[A person in a red jacket cross country skiing down a snow covered ski slope with a couple of people riding skis and walking on the side of the snowy mountain.]{%
      \includegraphics[clip, width=0.35\textwidth, height=0.5\textwidth]{figures/qualitative/10.jpeg}
      \hspace{0.03\textwidth}
      \includegraphics[clip, width=0.6\textwidth, height=0.5\textwidth]{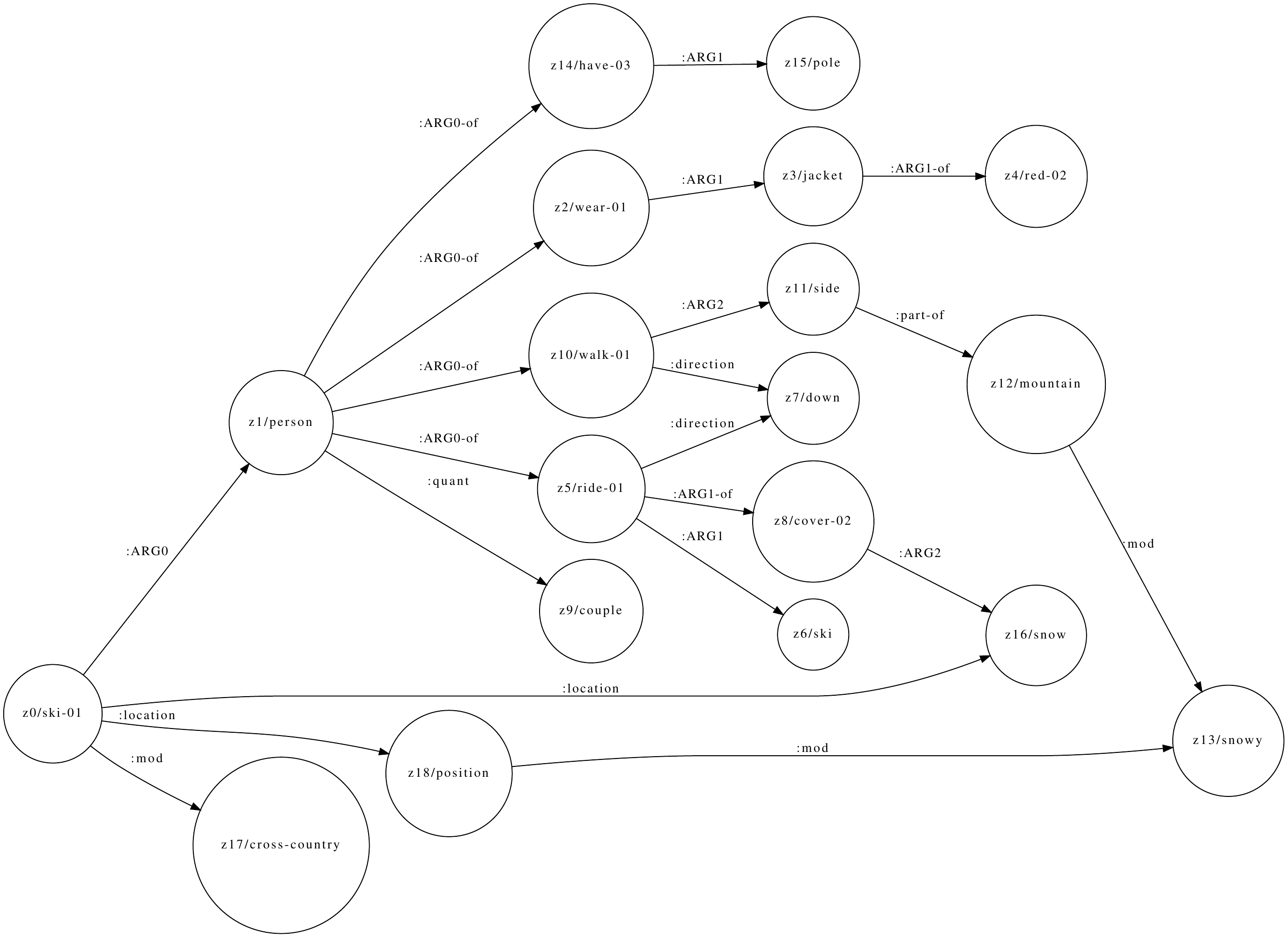}
    }
    \subfigure[A person in black shirt sitting at a table in a building with a plate of food with and smiling while having meal.]{%
      \includegraphics[clip, width=0.35\textwidth, height=0.5\textwidth]{figures/qualitative/11.jpeg}
      \hspace{0.03\textwidth}
      \includegraphics[clip, width=0.6\textwidth, height=0.5\textwidth]{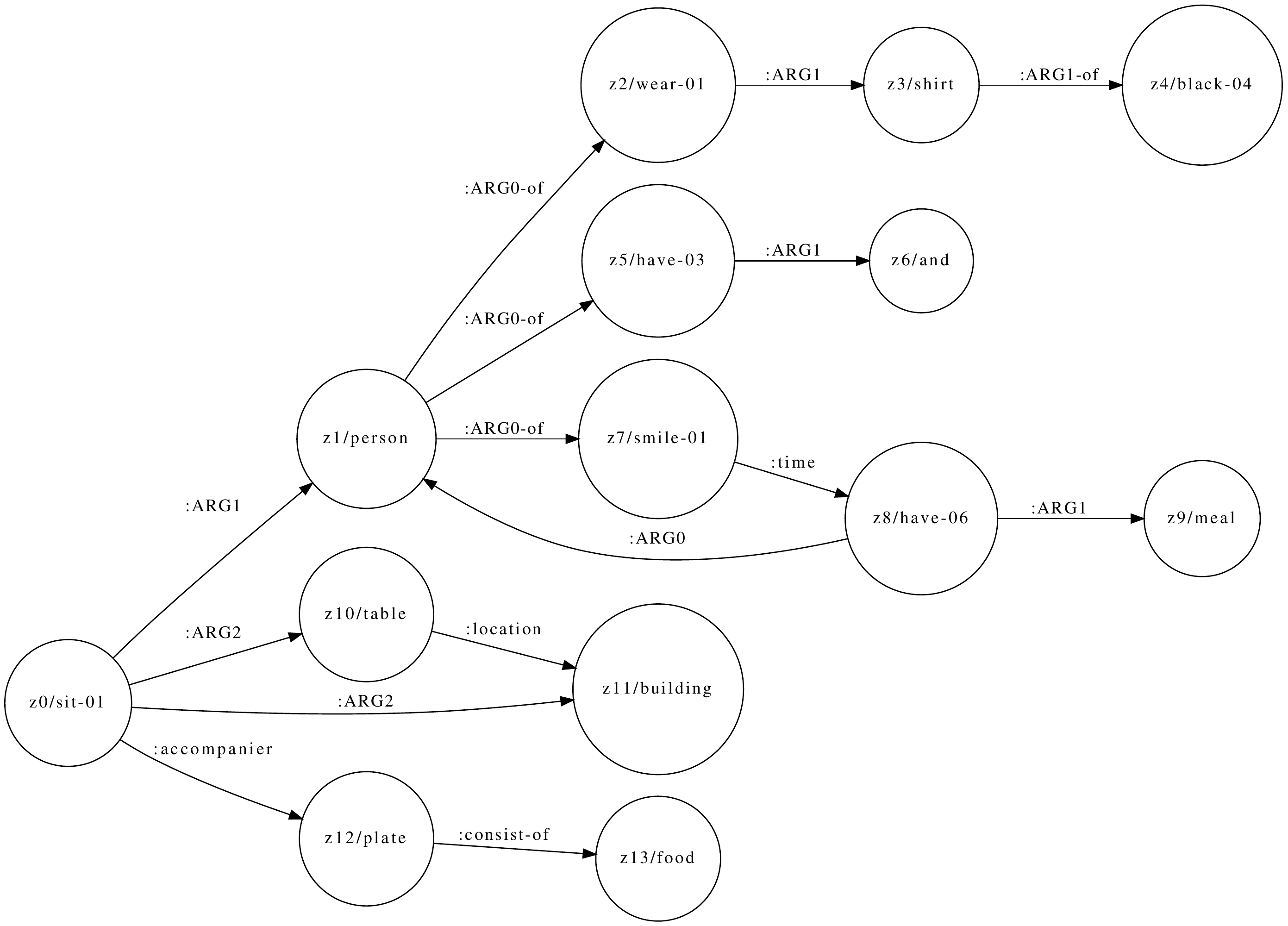}
    }
    \caption{(cont) Images used in Section~\ref{ss:qualitative}, along with their predicted AMRs and generated captions. Refer to Section~\ref{ss:qualitative} for more details.}
    \label{fig:qualitative_amrs_3}
\end{figure*}

\clearpage
Images used in this section (and the rest of the paper) are under a Creative Commons Attribution License 2.0 (\url{https://creativecommons.org/licenses/by/2.0/}). They are available at (by the order of their appearance in this section):
\begin{itemize}
    \item \url{http://farm6.staticflickr.com/5299/5465041730_3fe1246cae_z.jpg} and \url{http://cocodataset.org/#explore?id=505440}
    \item \url{http://farm6.staticflickr.com/5294/5461489420_1e4141517b_z.jpg} and \url{http://cocodataset.org/#explore?id=332654}
    \item \url{http://farm4.staticflickr.com/3719/9115013219_344a42ce47_z.jpg} and \url{http://cocodataset.org/#explore?id=329486}
    \item \url{http://farm4.staticflickr.com/3091/3187069218_162b55b720_z.jpg} and \url{http://cocodataset.org/#explore?id=569839}
    \item \url{http://farm3.staticflickr.com/2020/1932016761_934411ac16_z.jpg} and \url{http://cocodataset.org/#explore?id=5754}
    \item \url{http://farm4.staticflickr.com/3703/10047186866_e6b43fbd32_z.jpg} and \url{http://cocodataset.org/#explore?id=298443}
    \item \url{http://farm8.staticflickr.com/7170/6795850593_435a36bcd9_z.jpg} and \url{http://cocodataset.org/#explore?id=239235}
    \item \url{http://farm4.staticflickr.com/3786/9676804086_dbb624af5c_z.jpg} and \url{http://cocodataset.org/#explore?id=386559}
\end{itemize}

\end{document}